\documentclass[a4paper]{article}
\usepackage[T1]{fontenc}

\usepackage[utf8]{inputenc}
\usepackage{hyperref}
\hypersetup{
	unicode,
    hidelinks,
	breaklinks,
	pdfauthor={Kumar Abhishek, Aditi Jain, Ghassan Hamarneh},
	pdftitle={Investigating the Quality of DermaMNIST and Fitzpatrick17k Dermatological Image Datasets},
	pdfsubject={Investigating the Quality of DermaMNIST and Fitzpatrick17k Dermatological Image Datasets},
	pdfkeywords={dermatology, data quality in ML, deep learning},
	pdfproducer={LaTeX},
	pdfcreator={pdflatex}
}

\usepackage[pages=all, color=black, position={current page.south}, placement=bottom, scale=1, opacity=0, vshift=5mm]{background}
\usepackage[margin=1.05in]{geometry} %
\usepackage[sort,numbers,square]{natbib}
\usepackage{breakurl}

\makeatletter
\renewcommand*{\@fnsymbol}[1]{\ensuremath{\ifcase#1\or \dagger\or \ddagger\or \mathsection\or \mathparagraph\or \|\or **\or \dagger\dagger\or \ddagger\ddagger\else\@ctrerr\fi}}
\makeatother

\usepackage{authblk}    %

\usepackage{graphicx}
\usepackage{booktabs}
\usepackage{xspace}
\usepackage{caption}
\usepackage{subcaption}
\usepackage{url}
\usepackage{arydshln}
\usepackage{textcomp}
\usepackage{multirow}
\usepackage{dirtytalk}
\usepackage{soul}
\usepackage{bm}
\usepackage{amsmath}    %

\newcommand{\medmnist}{MedMNIST\xspace}
\newcommand{\dermamnist}{DermaMNIST\xspace}
\newcommand{\fitz}{Fitzpatrick\xspace}
\newcommand{\fitzdset}{Fitzpatrick17k\xspace}
\newcommand{\fitzdsetC}{Fitzpatrick17k-C\xspace}
\newcommand{\ham}{HAM10000\xspace}
\newcommand{\lowres}{$28 \times 28$\xspace}
\newcommand{\hires}{$224 \times 224$\xspace}
\newcommand{\hamcount}{10,015\xspace}
\newcommand{\fitzcount}{16,577\xspace}
\newcommand{\featdim}{960\xspace}
\newcommand{\dermaC}{DermaMNIST-C\xspace}
\newcommand{\dermaE}{DermaMNIST-E\xspace}
\newcommand{\pypackage}[1]{\texttt{#1}\xspace}
\newcommand{\eg}{e.g.,\xspace}
\newcommand{\ie}{i.e.,\xspace}
\newcommand{\diffF}{\hat{\mathcal{F}}^{\geq 1}}
\newcommand{\diffFStrict}{\hat{\mathcal{F}}^{> 1}}
\newcommand{\ka}[1]{{#1}}
\newcommand{\rev}[1]{{#1}}
\newcommand{\revnew}[1]{{#1}}
\newcommand{\saynew}[1]{``{#1}''}
\newcommand{\revnewer}[1]{{#1}}
\newcommand{\ghrepourl}{https://github.com/kakumarabhishek/Corrected-Skin-Image-Datasets}

\title{Investigating the Quality of DermaMNIST and Fitzpatrick17k Dermatological Image Datasets\thanks{\small{This manuscript has been accepted for publication in \emph{Nature Scientific Data} \textbf{12}, 196 (2025), {DOI:} \texttt{\url{https://doi.org/10.1038/s41597-025-04382-5}}. However, due to the publisher's human data policy, certain elements from this paper were omitted from the publisher's version of the manuscript, which have been reproduced here. Specifically, Figures 10 through 17 and the text supporting those visualizations is missing the publisher's version.}}}

\author[1,*]{Kumar Abhishek}
\author[2]{Aditi Jain}
\author[1]{Ghassan Hamarneh}
\affil[1]{School of Computing Science, Simon Fraser University, Canada}
\affil[2]{Department of Mathematics, Indian Institute Of Technology Delhi, India}
\affil[*]{Corresponding Author: Kumar Abhishek (\texttt{kabhishe@sfu.ca})}

\date{}

\begin{document}

\maketitle

\begin{abstract}
The remarkable progress of deep learning in dermatological tasks has brought us closer to achieving diagnostic accuracies comparable to those of human experts. However, while large datasets play a crucial role in the development of reliable deep neural network models, the quality of data therein and their correct usage are of paramount importance. Several factors can impact data quality, such as the presence of duplicates, data leakage across train-test partitions, mislabeled images, and the absence of a well-defined test partition. In this paper, we conduct meticulous analyses of three popular dermatological image datasets: \dermamnist, its source \ham, and \fitzdset, uncovering these data quality issues, measure the effects of these problems on the benchmark results, and propose corrections to the datasets. Besides ensuring the reproducibility of our analysis, by making our analysis pipeline and the accompanying code publicly available, we aim to encourage similar explorations and to facilitate the identification and addressing of potential data quality issues in other large datasets.
\end{abstract}

\section{Introduction}

Skin diseases are the most common reason for clinical consultations in studied populations~\cite{schofield2011skin}, affecting almost a third of the global population~\cite{hay2014global,flohr2021putting}. 
The 2013 Global Burden of Disease found skin diseases to be the fourth leading cause of nonfatal disabilities globally, accounting for 41.6 million Disability Adjusted Life Years and 39.0 million Years Lost due to Disability~\cite{karimkhani2017global}. In the USA alone, the healthcare cost of skin diseases was estimated to be \$75 billion in 2016~\cite{lim2017burden}.
Of all the skin diseases, skin cancer is a particularly concerning skin disease condition that merits special attention due to its potential seriousness.
With the increased incidence rates of skin cancer over the past decades~\cite{whocancer}, coupled with the projected decline in the ratio of dermatologists to populations~\cite{lim2017burden}, automated systems for dermatological diagnosis can be immensely valuable.

Advances in deep learning (DL)-based methods for dermatological tasks have produced models that are approaching the diagnostic accuracies of experts, some even mimicking clinical approaches of hierarchical~\cite{barata2019deep,benyahia2022hierarchical,yu2022skin} and differential~\cite{liu2020deep} diagnoses.
The data-driven nature of these DL methods implies that large and diverse datasets are needed to train accurate, robust, and generalizable models.
However, unlike natural computer vision datasets, medical image datasets are relatively smaller, primarily because of the large costs associated with image acquisition and annotation, legal, ethical, and privacy concerns~\cite{asgari2021deep}, and are more cost prohibitive to expand~\cite{kiryati2021dataset}.
This is also true for skin cancer image datasets~\cite{wen2022characteristics,mirikharaji2023survey}, where the surge in skin image analysis research over the past decade can be attributed in part to recent publicly available datasets, most notably the datasets and challenges of the International Skin Imaging Collaboration (ISIC) and the associated \ham~\cite{tschandl2018ham10000} and BCN2000~\cite{combalia2019bcn20000} datasets, which are primarily dermoscopic image datasets, as well as other clinical image datasets such as SD-198~\cite{sun2016benchmark}, SD-260~\cite{yang2019self}, derm7pt~\cite{kawahara2018seven}, and \fitzdset~\cite{groh2021evaluating}.

Although large data sets are important for the development of reliable models, the quality of the data therein and their correct use are equally important~\cite{norori2021addressing,sambasivan2021everyone,munappy2022data,budach2022effects}: \ka{low-quality data may result in inefficient training, inaccurate models that exhibit biases, poor generalizability and low robustness, and may negatively affect the interpretability of such models.
The  data quality} can be affected by several factors: mislabeled images, data leakage across training and evaluation partitions, 
\ka{the absence of a held-out test partition,}
etc.
The issue of data leakage, in particular, is in fact quite widespread, and a recent survey~\cite{kapoor2023leakage} of 17 fields spanning 294 articles, on topics ranging from medicine and bioinformatics to information technology operations and computer security, showed that ML adoption in all these fields suffers from data leakage. 
In an analysis of 10 popular natural computer vision, natural language, and audio datasets, Northcutt et al.~\cite{northcutt2021pervasive} estimated an average label error rate of at least 3.3\%. 
In medical image analysis domains, too, investigation into the use of machine learning best practices has found several instances of incorrect data partitioning and feature leakage between training and evaluation partitions. Oner et al.~\cite{oner2020training} showed that a peer-reviewed article published in Nature Medicine using DL for histopathology image analysis suffered from data leakage by using slide-level stratification for data partitioning instead of patient-level stratification. In a large-scale study, Bussola et al.~\cite{bussola2021ai} showed how DL models can exhibit considerably inflated performance measures when evaluated on datasets where histopathology image patches from the same subject are present in training and validation partitions. In mammography analysis, Samala et al.~\cite{samala2020hazards,samala2021risks} showed the risks of feature leakage between training and validation partitions and how this could lead to an overly optimistic performance on the validation partition, compared to a completely held out test partition.
Similar investigations have been conducted on the adverse effects of incorrect data partitioning on test performance in optical coherence tomography (OCT) image classification~\cite{tampu2022inflation}, brain magnetic resonance imaging (MRI) classification~\cite{yagis2021effect}, and longitudinal brain MRI analysis~\cite{rumala2023how}.

Specific to skin image analysis, 
our previous work~\cite{abhishek2020input}
showed that the popular ISIC Skin Lesion Segmentation Challenge datasets from 2016 through 2018 have considerable overlap among their training partitions, and that 706 images are present in all three datasets' training splits, a surprising discovery since ISIC 2016 only has 900 training images. Cassidy et al.~\cite{cassidy2022analysis} analyzed the ISIC Skin Lesion Diagnosis Challenge datasets from 2016 to 2020, and found overlap and duplicates across these datasets. They used a duplicate removal strategy to curate new clean training, validation, and testing sets. Vega et al.~\cite{vega2023analysis} found that a popular monkeypox skin image dataset, used in several peer-reviewed publications, contained \say{medically irrelevant images} and that models trained on these images did not necessarily rely on features underlying the diseases. Very recently, Groger et al.~\cite{groger2023reliable} carried out an analysis of six dermatology skin data sets (MED-NODE, PH2, DDI, derm7pt, PAD-UFES-20, and SD-128), detecting and removing near duplicates and \say{irrelevant samples} from them. However, all the datasets in their study were small, with 3 of 6 datasets (MED-NODE, PH2, DDI) consisting of less than 700 images and the largest (SD-128) containing 5,619 images.

In this paper, we look at 
\rev{three}
popular and large skin image analysis datasets: the \dermamnist dataset~\cite{yang2021medmnist,yang2023medmnist}
\rev{and its source \ham dataset~\cite{tschandl2018ham10000}}
(\hamcount{} images) and the \fitzdset dataset~\cite{groh2021evaluating} (\fitzcount{} images). We perform 
systematic analyses of both datasets and report instances of data duplication, data leakage across training and evaluation partitions, and data mislabeling in both datasets, fixing them where possible. Our corrected datasets and detailed analysis results are available online
at \url{https://derm.cs.sfu.ca/critique}.

\section{Results}
\label{sec:analysis}

\subsection{\dermamnist}
\label{subsec:medmnist_analysis}
Released as a 
biomedical imaging dataset equivalent
to the MNIST dataset of handwritten digits, \medmnist consists of
images from standardized biomedical imaging datasets resized to MNIST-like \lowres resolution. Despite being a fairly new dataset, it has been quite popular 
\rev{(\revnewer{995} citations as of \revnewer{November}, 2024: \revnewer{622} citations of [a more recent] 2023 paper~\cite{yang2023medmnist}, and \revnewer{373} of [an older] 2021 version of the paper~\cite{yang2021medmnist}).}
The dermatological subset of \medmnist, \dermamnist, contains resized images from the popular \say{Human Against Machine with 10000 training images} (\ham) dataset~\cite{tschandl2018ham10000}. \ham contains \hamcount dermoscopic images of pigmented skin lesions
collected from patients at two study sites in Australia and Austria, with their diagnoses confirmed by either histopathology, confocal microscopy, clinical follow-up visits, or expert consensus. The 7 disease labels in the dataset cover 95\% of the lesions encountered in clinical practice~\cite{tschandl2018ham10000}. Because of these meritorious properties, \ham is a good candidate 
dataset for dermatological analysis, as aimed for with \dermamnist.
\rev{While the \say{\textit{lightweight}} nature of \dermamnist due to its \say{\textit{small size}} is appealing for its adoption in machine learning for biomedical imaging~\cite{yang2023medmnist}, the low spatial resolution (\lowres) does not capture sufficient morphological structures of skin lesions compared to the source dataset \ham.
Despite this, \dermamnist has been used for a wide variety of applications in peer-reviewed publications: 
semi- and self-supervised learning~\cite{sportisse2023are,kang2023label,chen2023knowledge}, 
federated learning~\cite{lu2023federated,zhu2022federated,kim2023re}, 
privacy-preserving learning~\cite{tang2023differentially,lee2023hetal}, 
neural architecture search~\cite{xu2023detection,wang2024mednas}, 
adversarially robust learning~\cite{xu2022medrdf,ahmed2022failure}, 
data augmentation~\cite{yang2022adversarial,abhishek2024multi}, 
generative modeling~\cite{wang2024autoencoder}, 
model interpretability~\cite{nguyen2023mononet},
AutoML~\cite{kumar2023extending},
\revnewer{
active learning~\cite{das2023accelerating},
quantum vision transformers~\cite{cherrat2024quantum}, and
biomedical vision-language foundation models~\cite{lin2023pmc,zhang2024generalist,zhu2024unimed},
as well as derivative benchmark datasets~\cite{di2024medmnist,doerrich2024rethinking}.}}
However, as we investigate below, the resulting \dermamnist and its benchmarks suffer from serious flaws.

\subsubsection{Data Leakage}

\begin{figure}[!ht]
     \centering
     \begin{subfigure}[b]{0.6\textwidth}
         \centering
         \includegraphics[width=\textwidth]{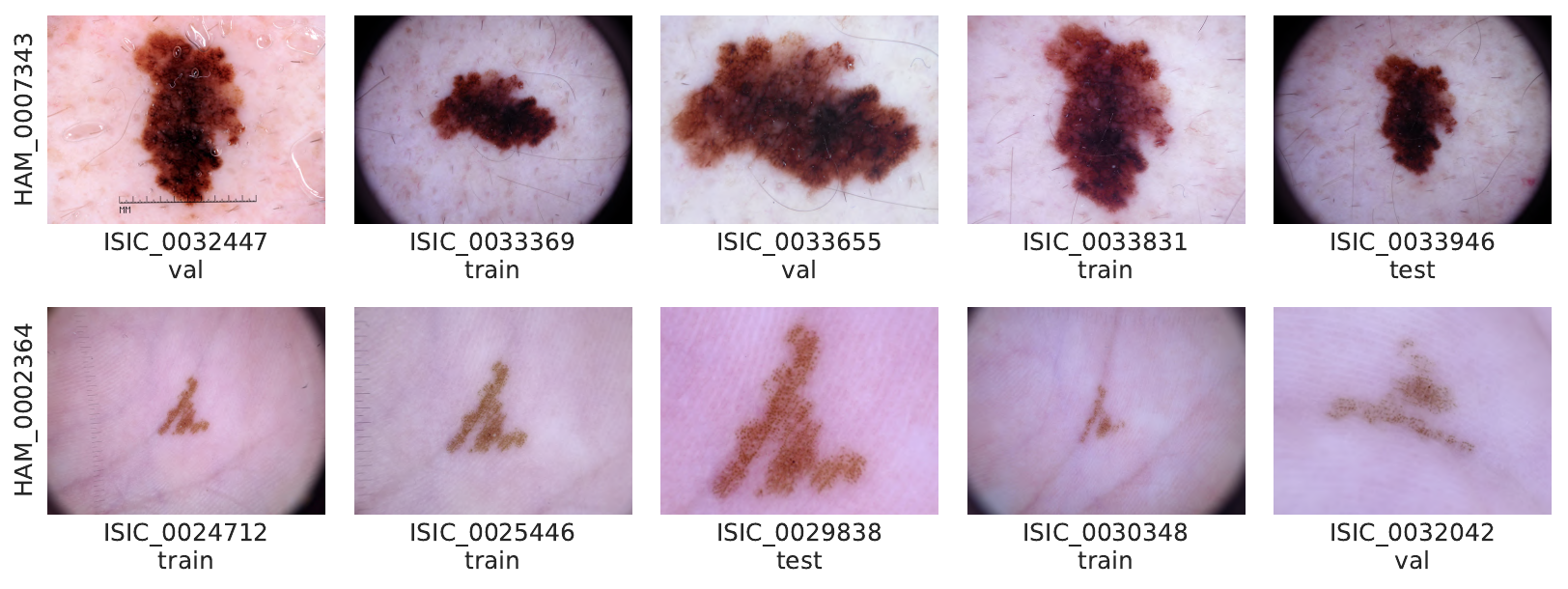}
         \caption{Data leakage: Images of the same lesion are present in multiple partitions (train, valid, and test).
         }
     \end{subfigure}
     \quad
     \begin{subfigure}[b]{0.37\textwidth}
         \centering
         \includegraphics[width=\textwidth]{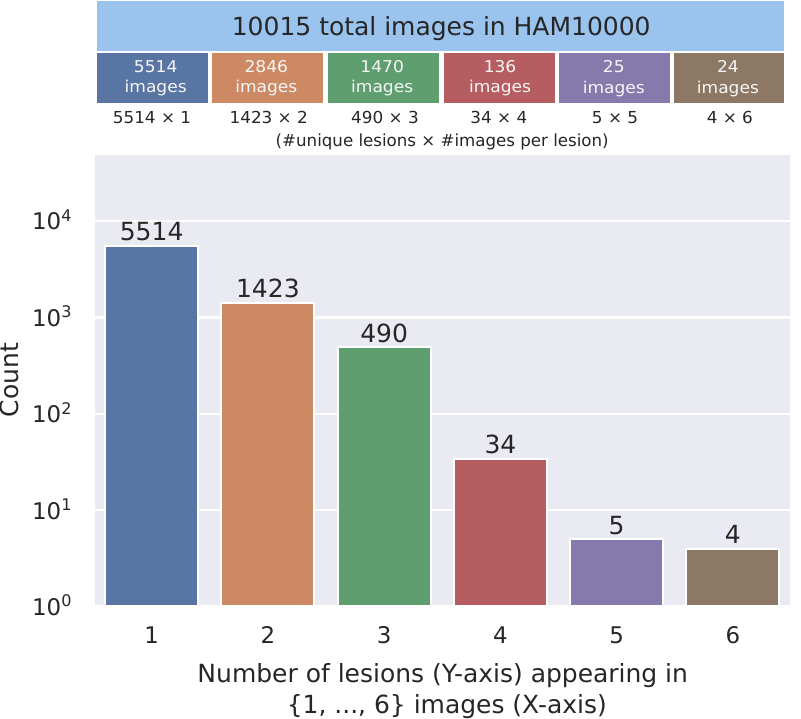}
         \caption{\ka{Breakdown of \ham by the number of unique lesions.}}
     \end{subfigure}
     \\   
     \begin{subfigure}[b]{0.95\textwidth}
         \centering
         \includegraphics[width=\textwidth]{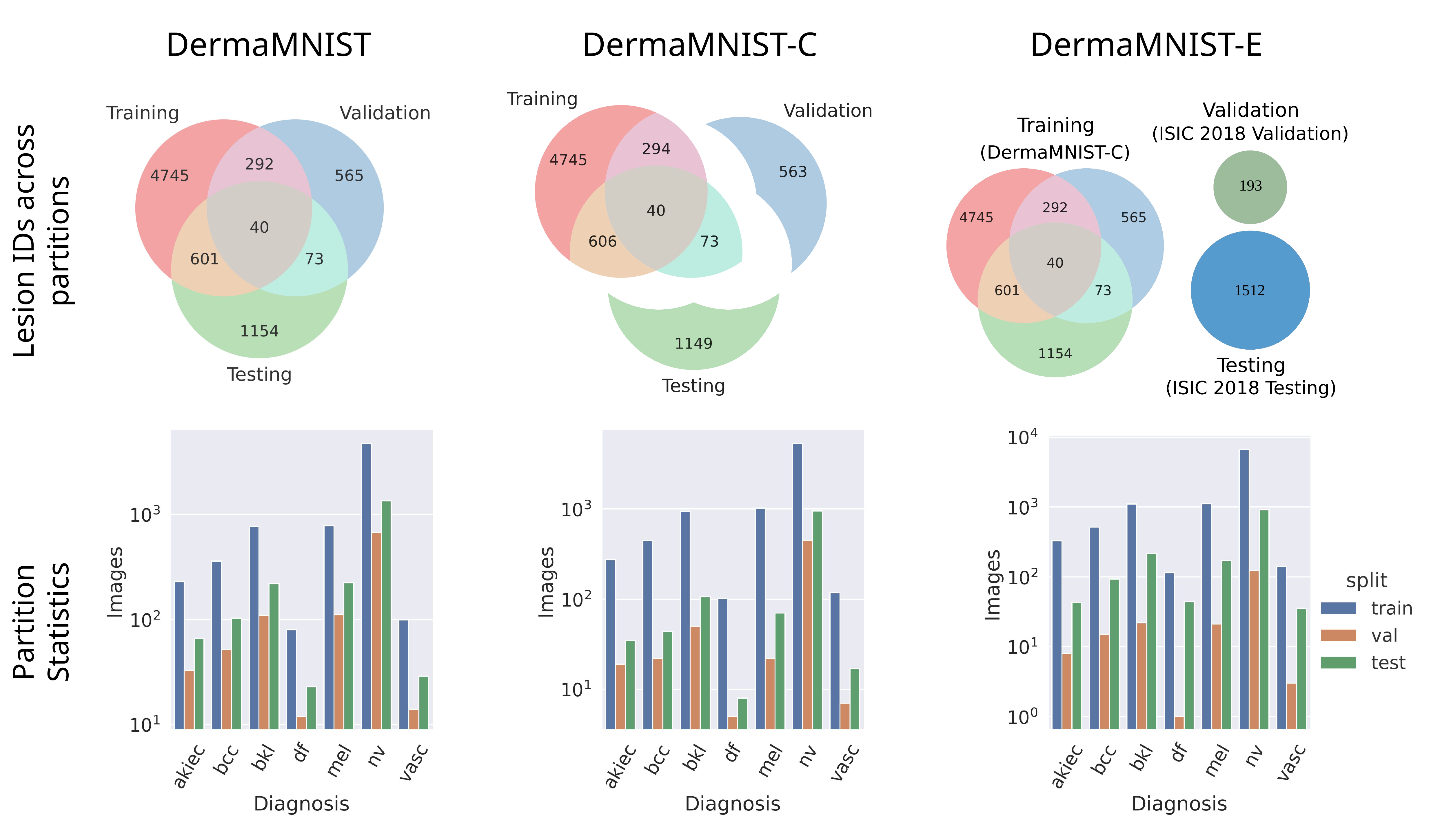}
         \caption{
         \rev{
         \textbf{Top row:} Venn diagrams visualize how unique \texttt{lesion IDs} are distributed across partitions of \dermamnist, \dermaC, and \dermaE. Note that the overlap between partitions in \dermamnist is addressed by \dermaC and \dermaE. \textbf{Bottom row:} Distributions of diagnosis labels across partitions in the three datasets.
         }
         }
     \end{subfigure}

     \caption{\dermamnist analyses: \textbf{(a, b)} show instances of and reasons for the data leakage, and \textbf{(c)} visualizes how the three datasets: \dermamnist, \dermaC, and \dermaE differ in their partition composition, yet have similarly proportionate diagnosis distributions.    
         \revnew{Images from \dermamnist are licensed under CC BY-NC 4.0~\cite{yang2021medmnist,yang2023medmnist}.}  Best viewed online.
     }
     \label{fig:dermamnist}
\end{figure}

\begin{figure}[!ht]
     \centering
     \begin{subfigure}[b]{\textwidth}
         \centering
         \includegraphics[width=\textwidth]{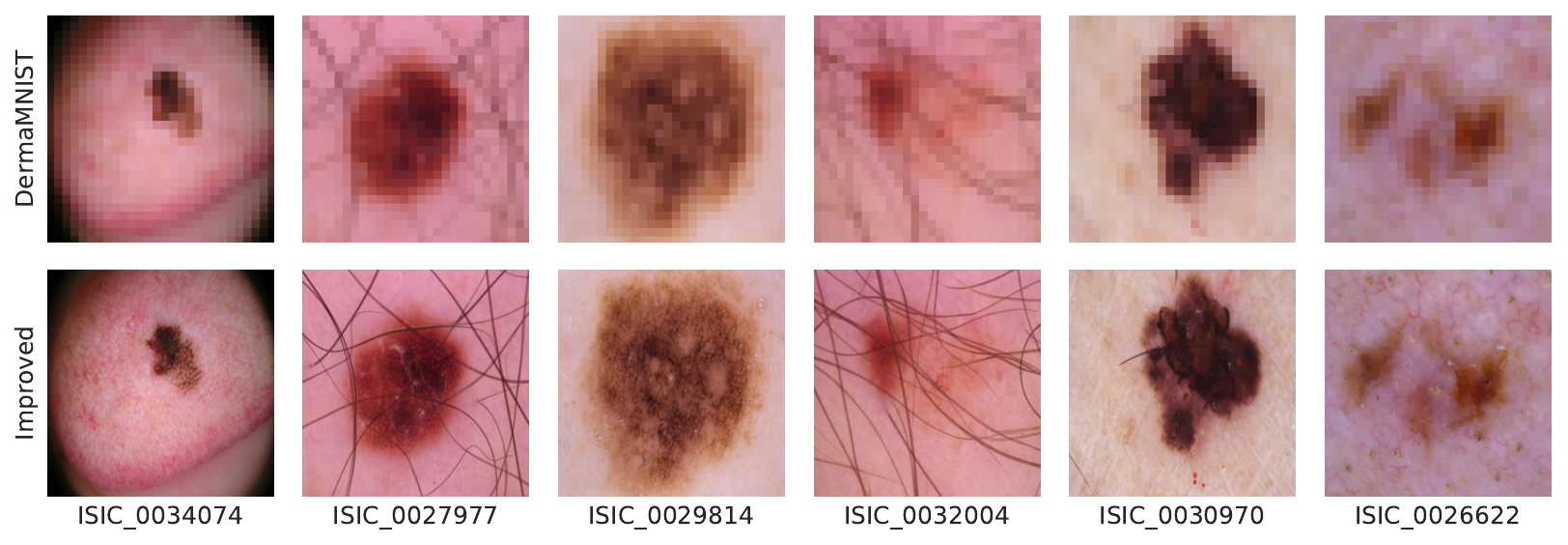}
     \end{subfigure}

     \caption{Visualizing how \dermamnist's incorrect resizing operation leads to loss of information.
     \dermamnist's approach (\textbf{top row}) to generating \hires images results in visibly pixelated images. Our approach (\textbf{bottom row}), used for both \dermaC and \dermaE, retains much more detailed information. \revnew{Images from \dermamnist are licensed under CC BY-NC 4.0~\cite{yang2021medmnist,yang2023medmnist}.}
     Best viewed online.
     }
     \label{fig:dermamnist_part2}
\end{figure}

A caveat of \ham, despite its rather large size, is that it contains multiple images of the same lesion captured either from different viewing angles or at different magnification levels (Fig.~\ref{fig:dermamnist} (a)),
\ie the number of lesions with unique \texttt{lesion ID}s (\texttt{HAM\_xxx}) is smaller than the number of images with unique \texttt{image ID}s (\texttt{ISIC\_xxx}).
We visualize the frequency counts of lesions and how many images of the same lesion are present in \ham in Fig.~\ref{fig:dermamnist} (b) and observe that the \hamcount{} images are in fact derived from only 7,470 unique lesions, and
1,956 of these \texttt{lesion ID}s ($\sim$26.18\%) contains 2 or more images: 1,423 lesions have 2 images, 490 lesions have 3 images, 34 lesions have 4 images, 5 lesions have 5 images, and 6 lesions have 4 images each. Unfortunately, this was not accounted for when preparing train-valid-test splits for \dermamnist. The \dermamnist dataset is released as a pre-processed NumPy array, so while the dataset does not contain filenames of the images, we confirmed this by contacting the authors to obtain the exact training-validation-testing split filenames. Therefore, there is considerable data leakage of images of the same lesion across partitions. 
Fig.~\ref{fig:dermamnist} (a) shows 2 examples
where images of the same lesion are present in the training, validation, and testing partitions.
This issue is quite pervasive across \dermamnist, and our analysis found the following overlaps across partitions: train-test: 886 images (641 lesions), train-valid: 440 images (332 lesions); valid-test: 128 images (113 lesions); train-valid-test: 51 images (40 lesions) (Fig.~\ref{fig:dermamnist} (c)).
Such data leakage naturally raises concerns on the reliability of the \dermamnist benchmarks and related studies.

We 
correct this data leakage by simply moving all 
images of a \texttt{lesion ID} present in the train partition from valid and test partitions back to the train partition.
\ka{We choose to do this (i.e., moving the images of a lesion to the training set) instead of removing the images completely, 
to ensure that images of the same lesion are in one partition
while also not discarding any images.}
Although this has the undesirable side effect of increasing the training partition size at the cost of reduced validation and testing partition sizes, it fixes the data leakage issue by ensuring there is no overlap across partitions. 

\begin{figure}[!ht]
\centering
\includegraphics[width=0.9\textwidth]{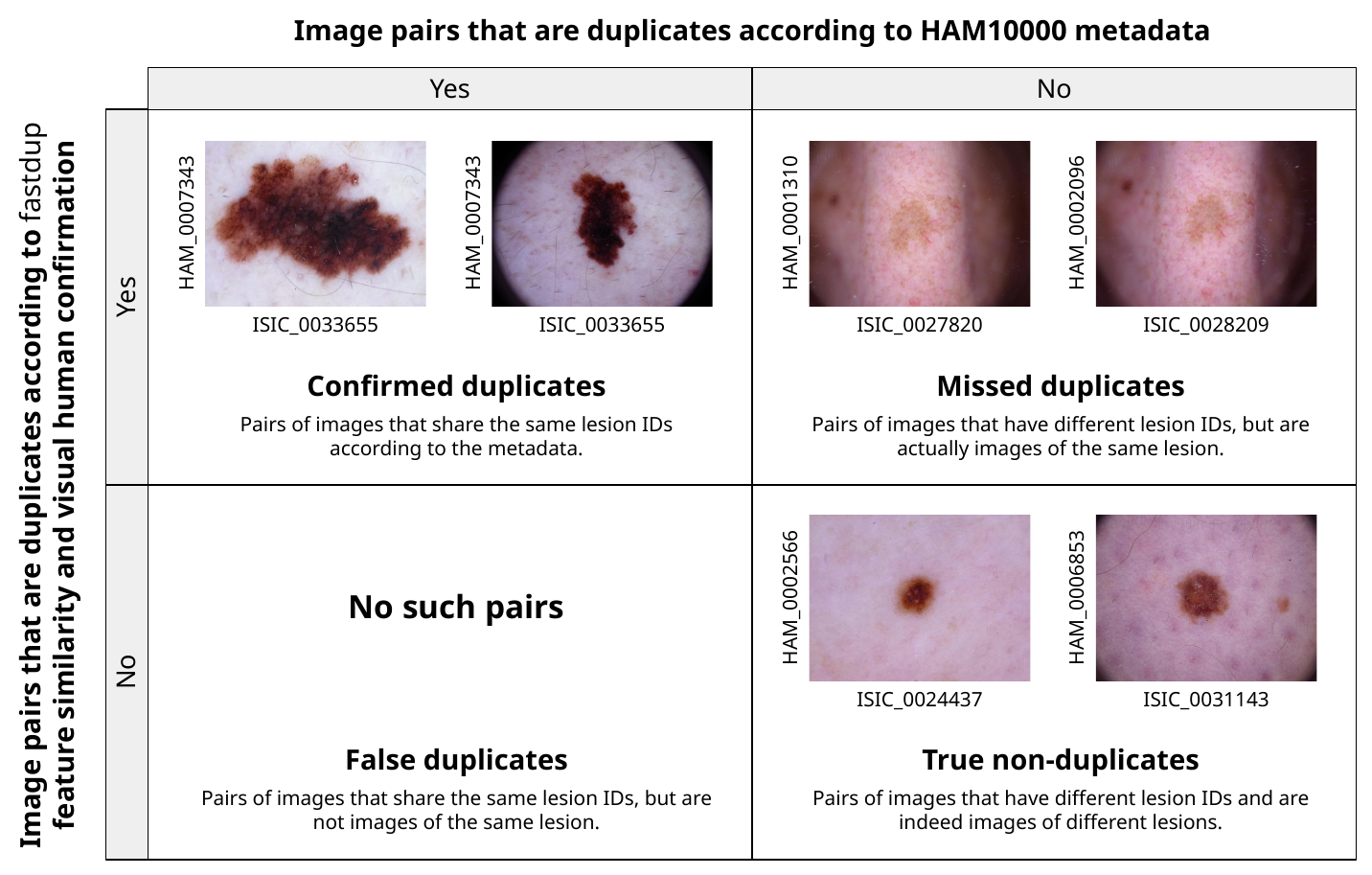}
\caption{
\rev{Visualizing the four scenarios that a pair of images from \ham can be assigned to in duplicate detection, based on the metadata and the \pypackage{fastdup}-based duplicate detection followed by manual review. \saynew{Confirmed duplicates}, as the name suggests, are pairs that are images of the same lesion, indicated by the same \texttt{lesion ID}s in the metadata. Similarly, \saynew{True non-duplicates} are pairs of images that belong to different lesions. \saynew{Missed duplicates} refer to image pairs that have differing \texttt{lesion ID}s according to the metadata, but their high visual similarity (measured by cosine similarity of their image embeddings) followed by manual review confirms that these are indeed images of the same lesion, and were therefore `missed` by the metadata. Finally, \saynew{False duplicates} refer to pairs where images share the same \texttt{lesion ID}s but do not belong to the same lesion. In our analysis, we did not find any instances of \saynew{False duplicates} in \ham.
For all these sample images, the \texttt{image ID}s and the \texttt{lesion ID}s are along the horizontal and the vertical axis, respectively.
\revnew{Images from \ham are licensed under CC BY-NC 4.0~\cite{tschandl2018ham10000}.}
}
}
\label{fig:ham10000_confusion_matrix}
\end{figure}

\rev{Next, we examine the accuracy of \ham's metadata.
While the images in \ham and their labels have been collected from clinical sites and the dataset itself has been widely adopted for online challenges~\cite{codella2019skin,isic2019international} and human-in-the-loop evaluations~\cite{tschandl2020human,barata2023reinforcement}, we found that some images with different \texttt{lesion ID}s are in fact duplicates and should have been assigned the same \texttt{lesion ID}. 
Therefore, for a systematic analysis, we use \pypackage{fastdup}~\cite{fastdup_cite}
an open-source Python library for analyzing visual datasets at scale, and calculate inter-image embedding similarity scores $\mathcal{S} (x_i, x_j)$ for all
$\binom{\hamcount{}}{2}$
pairs of images \rev{$(x_i, x_j)$}.
Fig.~\ref{fig:ham10000_confusion_matrix} visualizes, as a confusion matrix, the following four scenarios that are possible when comparing duplication checks based on the metadata with duplicates detected using \pypackage{fastdup} followed by visual human confirmation:
\begin{itemize}
    \item \say{Confirmed duplicates}: image pairs where both images share the same \texttt{lesion ID}s in the metadata and are indeed images of the same lesion.
    \item \say{True non-duplicates}: image pairs where both images differ in their \texttt{lesion ID}s and are images of different lesions.
    \item \say{Missed duplicates}: image pairs where both images differ in their \texttt{lesion ID}s but actually belong to the same lesion.
    \item \say{False duplicates}: image pairs where both images share the same \texttt{lesion ID}s in but actually belong to the same lesion.
\end{itemize}
}

\rev{The image pairs that lie along this confusion matrix's diagonal, i.e., \say{confirmed duplicates} and \say{true non-duplicates}, are those where the metadata agrees with our analysis, and the possible errors in metadata arise out of the other two scenarios.}

\begin{figure}[!ht]
\centering
\includegraphics[width=0.85\textwidth]{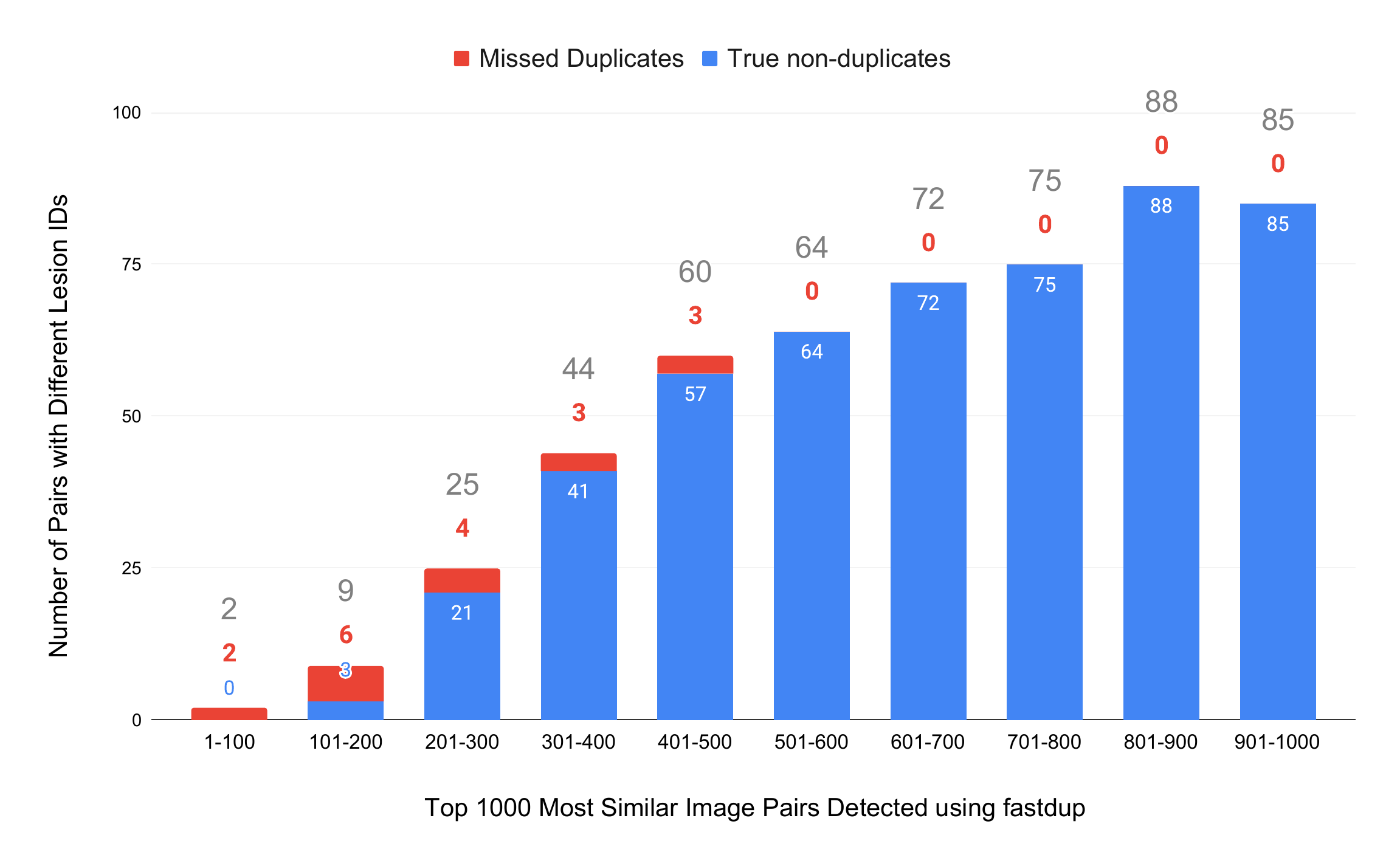}
\caption{
\rev{Analysis of the top 1,000 most similar pairs in \ham detected by \pypackage{fastdup}: in intervals of 100 images, we calculate how many of these 100 purported duplicate image pairs are not already present in the \ham metadata, and manually review those to detect which of these are \saynew{Missed duplicates} (i.e., pairs where the two images have different \texttt{lesion ID}s, but are actually images of the same lesion; Fig.~\ref{fig:ham10000_confusion_matrix}) and those that are \saynew{True non-duplicates} (i.e., pairs where the two images have different \texttt{lesion ID}s but are indeed images of different lesions; Fig.~\ref{fig:ham10000_confusion_matrix}). For example, looking at the 301--400 range, we find that from the 301\textsuperscript{st} to the 400\textsuperscript{th} most similar image pairs detected by \pypackage{fastdup}, 44 pairs contained images that did not belong to the same \texttt{lesion ID} according to the \ham metadata. Of these 44 pairs, manual inspection revealed 3 pairs to be newly discovered \saynew{Confirmed Duplicates}, whereas the remaining 41 pairs were images of different lesion and were therefore \saynew{False Positives}. There were 18 confirmed duplicate image pairs detected in \ham and they have been visualized in Fig.~\ref{fig:ham10000_most_similar}.}
}
\label{fig:ham10000_duplicates_bar_chart}
\end{figure}

\begin{figure}[!ht]
\centering
\includegraphics[width=\textwidth]{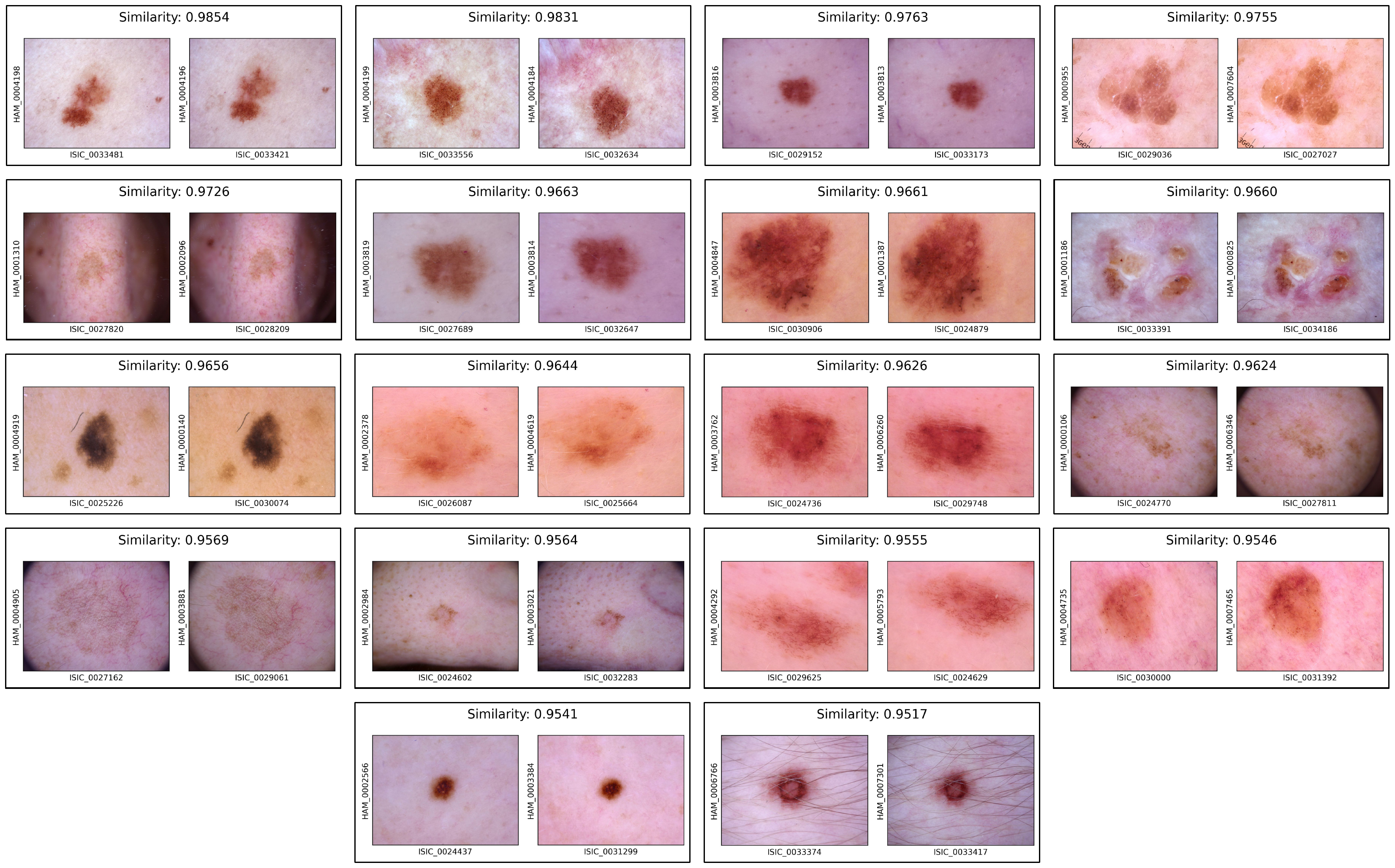}
\caption{
\rev{
Visualizing the 18 \saynew{Missed duplicates} (Fig.~\ref{fig:ham10000_confusion_matrix}) in \ham obtained through the analysis of the top 1,000 most similar image pairs (Fig.~\ref{fig:ham10000_duplicates_bar_chart}). These 18 pairs of images (\texttt{image ID}s along the horizontal axis) should belong to different lesions (\texttt{lesion ID}s along the vertical axis) according to the metadata, but manual review shows that both images in these pairs belong to the same lesions, and are thus, duplicate image pairs. \revnew{Images from \ham are licensed under CC BY-NC 4.0~\cite{tschandl2018ham10000}.}}
}
\label{fig:ham10000_most_similar}
\end{figure}

\begin{figure}[!ht]
\centering
\includegraphics[width=0.75\textwidth]{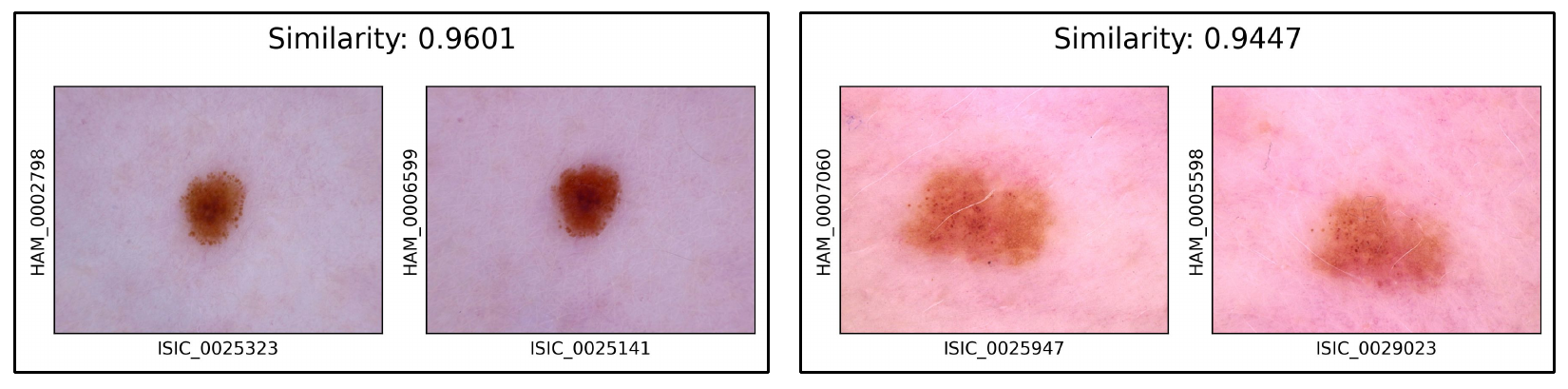}
\caption{
\rev{Two image pairs (\texttt{image ID}s along the horizontal axis) that have a high visual similarity but belong to different \texttt{lesion ID}s (vertical axis) according to the \ham metadata. Upon closer inspection, the images are near duplicates but exhibit inconspicuous differences, and are possibly images of the same lesion acquired at different times. \revnew{Images from \ham are licensed under CC BY-NC 4.0~\cite{tschandl2018ham10000}.}}
}
\label{fig:ham10000_maybe_different_times}
\end{figure}

\begin{figure}[!ht]
\centering
\includegraphics[width=\textwidth]{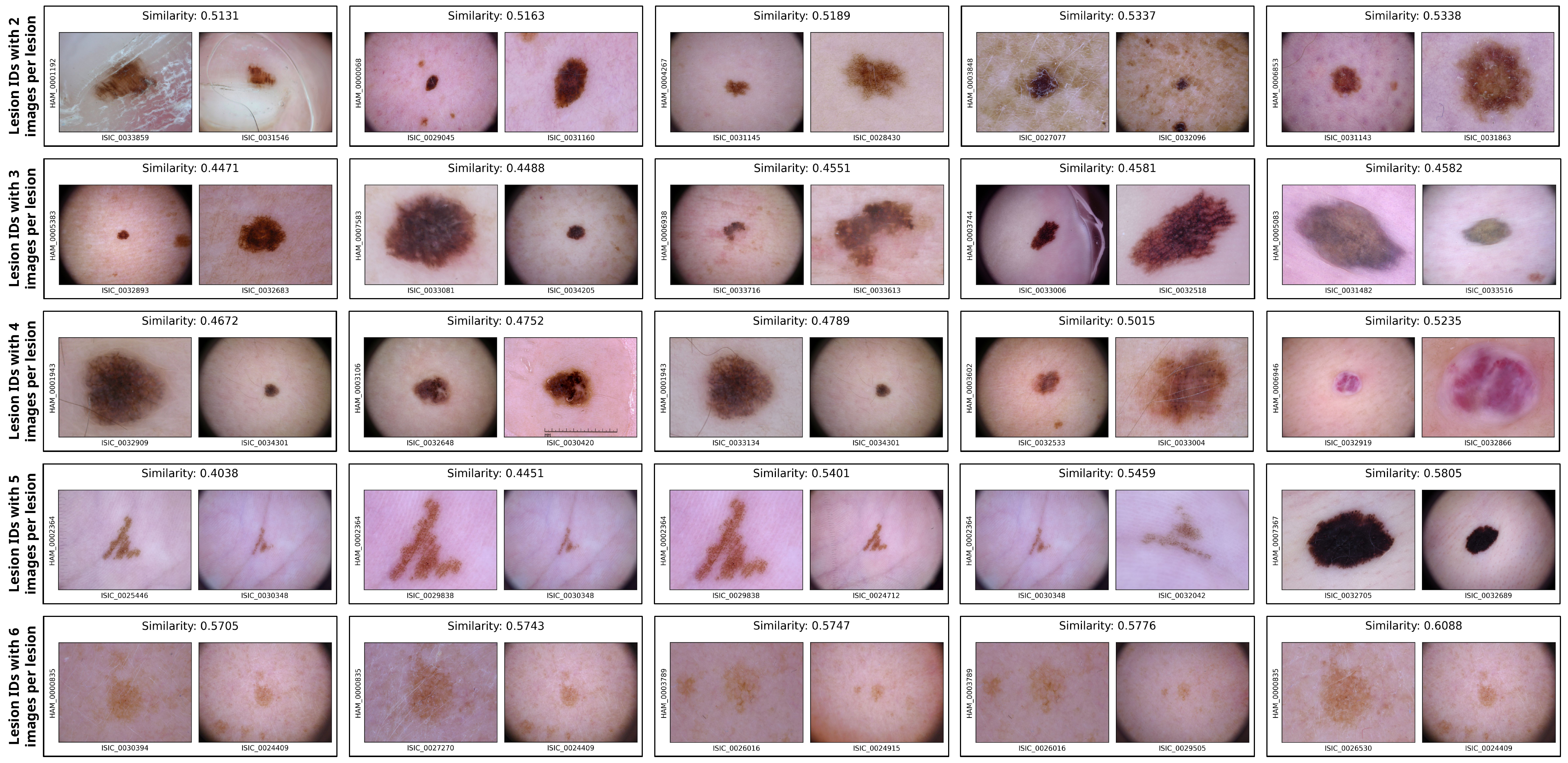}
\caption{
\rev{
Visualizing the least similar image pairs that belong to the same lesion in \ham to check for \saynew{False duplicates} (Fig.~\ref{fig:ham10000_confusion_matrix}). For each \texttt{lesion ID} with \{2, 3, 4, 5, 6\} images per lesion (Fig.\ref{fig:dermamnist} (b)), we look at pairs of images (\texttt{image ID}s along the horizontal axis) that belong to the same lesion (\texttt{lesion ID} along the vertical axis) but have the lowest similarity scores calculated using \pypackage{fastdup}. We do this to detect whether images with mislabeled \texttt{lesion ID}s, since images of two different lesions that have been assigned the same \texttt{lesion ID} will be dissimilar and therefore have a low similarity score. In each row, we visualize the 5 least similar image pairs that share a \texttt{lesion ID}, and do so for lesions that have \{2, 3, 4, 5, 6\} images (Fig.\ref{fig:dermamnist} (b)). We observe that all the image pairs indeed belong to the same lesion, and the low similarity score values can be easily explained by different zoom levels and/or geometric transformations (e.g., rotation and flipping).
\revnew{Images from \ham are licensed under CC BY-NC 4.0~\cite{tschandl2018ham10000}.}
}
}
\label{fig:ham10000_least_similar}
\end{figure}

\rev{For detecting errors of the first kind, i.e., 
\say{missed duplicates}, 
we analyze the top 1,000 most similar image pairs, measured by similarity of the image embeddings. 
We look at these 1,000 pairs in intervals of 100 pairs, ordered by decreasing inter-image similarity. For the 100 image pairs in each interval, we look up the \ham metadata for both images in a pair to exclude 
\say{confirmed duplicates}, since these duplicates are already accounted for in the metadata. For the remaining pairs, we manually review them to detect which, if any, of these are \say{missed duplicates}
and 
\say{true non-duplicates}.
We visualize these counts in Fig.~\ref{fig:ham10000_duplicates_bar_chart}. 
Of the 1,000 most similar image pairs in \ham, we discover 18 \say{missed duplicates} image pairs that were not accounted for in the metadata, which are visualized in Fig.~\ref{fig:ham10000_most_similar}. Moreover, the fraction of \say{true non-duplicates} keeps monotonically increasing as we look at intervals of 100 image pairs, going from $0\%$ (0 \say{true non-duplicates} out of 2 duplicate pairs) in the top 100 most similar pairs to $100\%$ (64 \say{true non-duplicates} out of 64 duplicate pairs) in the 501\textsuperscript{st} to the 600\textsuperscript{th} most similar pairs. This may be explained by the narrow field of view in dermoscopic images (\ham contains dermoscopic images), allowing for fewer visual cues for accurate detection of duplicates. Additionally, the lack of any \say{missed duplicates} after the top 500 most similar pairs (Fig.~\ref{fig:ham10000_duplicates_bar_chart}) means it is highly unlikely that, apart from the 18 duplicate pairs discovered (Fig.~\ref{fig:ham10000_most_similar}), there are other undetected duplicate image pairs in \ham.}

\rev{Fig.~\ref{fig:ham10000_maybe_different_times} shows an interesting discovery in the manual review of the highly similar image pairs. We found two instances of image pairs with high similarity scores, but manual inspection confirmed that the lesions had minor morphological differences. \ham contains images acquired during follow-up clinical visits~\cite{tschandl2018ham10000}, and it is possible that these highly similar image pairs are images of the same lesion acquired at different time durations.}

\rev{Next, we check of errors of the second kind, i.e., \say{false duplicates}.
For all the \texttt{lesion ID}s that have more than 1 image per lesion, we measure image similarity between images that belong to the same lesion and manually review the 5 least similar image pairs, as visualized in Fig.~\ref{fig:ham10000_least_similar}. We find that there are no errors, meaning that all image pairs, despite their low similarities, indeed belong to the same lesion, and that visual dissimilarities can be attributed to one or more of: zoom and crop levels, rotation, flipping, and artifacts such as gel bubbles and rulers.}

\rev{Of these 18 newly discovered duplicate image pairs in \ham, 7 pairs leak across partitions: train-test: 5 images (5 lesions) and train-valid: 2 images (2 lesions). We correct this data leakage by moving both images in each of these 7 pairs to the train partition.}

We name this \say{corrected} dataset version as \textbf{\dermaC}. Furthermore, the relative diagnosis-wise distribution of \dermaC across partitions is quite similar to that of the original \dermamnist (Fig.~\ref{fig:dermamnist} (c)). We benchmark \dermaC using \dermamnist's publicly available disease classification codebase, 
repeating all experiments 3 times for robustness, 
and the results 
are presented in Table~\ref{tab:dermamnist}. 

\subsubsection{Results on \hires resolution}
\dermamnist is created by resizing images from \ham's original $600 \times 450$ spatial resolution to MNIST-like \lowres resolution using (bi)cubic spline interpolation. However, for their classification benchmark experiments on the \hires resolution, instead of resizing the original images to \hires, the authors\cite{yang2021medmnist,yang2023medmnist} upsample the low\rev{-}resolution \lowres to obtain the \hires using nearest neighbor 
interpolation
(\ka{\say{\textit{224 (resized from 28)}}; verifiable through their source code}~\cite{dermamnist_sampling_error}).
This, unsurprisingly, leads to quite 
blurry images, given the unrecoverable information lost when downsampling from the original $600 \times 450$ to \lowres
and leads to significant loss of detail (e.g., dermoscopic structures and artifacts) in the images used to train models on \hires images 
(Fig.~\ref{fig:dermamnist_part2}).
Our approach of directly downsampling from the original high\rev{-}resolution to \hires to create \dermaC and \dermaE, used when reporting results in Table~\ref{tab:dermamnist}, yields conspicuously more detailed outputs 
(Fig.~\ref{fig:dermamnist_part2}).
\begin{table}[!ht]
\centering
\resizebox{0.9\textwidth}{!}{%
\setlength{\tabcolsep}{0.7em}
\def\arraystretch{1.25}
\begin{tabular}{@{}clcc:cc@{}}
\toprule
\multicolumn{2}{c}{\multirow{3}{*}{\textbf{Dataset}}} & \multicolumn{4}{c}{\textbf{Method}} \\ \cdashline{3-6} 
\multicolumn{2}{c}{} & \multicolumn{2}{c}{\textbf{ResNet-18}} & \multicolumn{2}{c}{\textbf{ResNet-50}} \\ 
\multicolumn{2}{c}{} & \multicolumn{1}{c}{AUC} & \multicolumn{1}{c}{ACC} & \multicolumn{1}{c}{AUC} & ACC \\ \midrule
\multicolumn{1}{c}{\multirow{2}{*}{\textbf{\dermamnist}}} & Original-28 & \multicolumn{1}{c}{0.917} & \multicolumn{1}{c}{0.735} & \multicolumn{1}{c}{0.913} & 0.735 \\ 
\multicolumn{1}{c}{} & Original-224 & \multicolumn{1}{c}{0.920} & \multicolumn{1}{c}{0.754} & \multicolumn{1}{c}{0.912} & 0.731 \\ \midrule

\multicolumn{1}{c}{\multirow{2}{*}{\textbf{\dermaC}}} & \lowres & \multicolumn{1}{c}{\rev{0.946} $\pm$ \rev{0.003}} & \multicolumn{1}{c}{0.825 $\pm$ \rev{0.006}} & \multicolumn{1}{c}{\rev{0.943} $\pm$ \rev{0.012}} & \rev{0.831} $\pm$ \rev{0.019} \\ 
\multicolumn{1}{c}{} & \hires & \multicolumn{1}{c}{\rev{0.953} $\pm$ \rev{0.001}} & \multicolumn{1}{c}{\rev{0.847} $\pm$ \rev{0.010}} & \multicolumn{1}{c}{\rev{0.948} $\pm$ \rev{0.001}} & \rev{0.847} $\pm$ 0.005 \\ \midrule

\multicolumn{1}{c}{\multirow{2}{*}{\textbf{\dermaE}}} & \lowres & \multicolumn{1}{c}{0.892 $\pm$ 0.006} & \multicolumn{1}{c}{0.676 $\pm$ \rev{0.009}} & \multicolumn{1}{c}{0.888 $\pm$ 0.003} & 0.674 $\pm$ 0.009 \\ 
\multicolumn{1}{c}{} & \hires & \multicolumn{1}{c}{0.896 $\pm$ 0.010} & \multicolumn{1}{c}{0.685 $\pm$ \rev{0.000}} & \multicolumn{1}{c}{0.901 $\pm$ 0.008} & 0.682 $\pm$ 0.003 \\ \bottomrule
\end{tabular}%
}
\caption{Benchmark results (3 repeated runs; mean $\pm$ std. dev.) of \dermamnist and the 2 proposed versions: \dermaC and \dermaE. The metrics being reported are the overall area under the ROC curve (AUC) and the overall accuracy (ACC).}
\label{tab:dermamnist}
\end{table}

\subsubsection{Extending \dermamnist:}

Finally, although \dermaC is a good light-weight dataset of choice for evaluating machine learning models on dermatological tasks and for educational purposes, as \medmnist intended to be, the quantitative results
of \dermaC (and \dermamnist for that matter)
perhaps paint a deceptively optimistic picture of the state of automated dermatological diagnosis models. 
We propose a more challenging extension of \dermamnist named 
\textbf{\dermaE}. 
The original \dermamnist and its corrected version \dermaC are based on \ham, which was used as the training partition for the ISIC Challenge 2018. However, the ISIC Challenge 2018 
had, apart from the \hamcount{} training images from \ham, 
separate validation and testing partitions containing 193 and 1,512 images, respectively. 
Therefore, we create \dermaE with \dermamnist as training set and ISIC 2018 validation and test partitions as validation and testing sets, respectively. \rev{Although the official testing partition of ISIC 2018 contained 1,512 images, we remove one image known as the \say{easter egg} (\texttt{ISIC\_0035068})~\cite{barata2023reinforcement}, resulting in a total of 1,511 images.}
While the resulting diagnosis distribution across partitions for \dermaE is similar to that of \dermaC and \dermamnist (Fig.~\ref{fig:dermamnist} (c)), our benchmark results on \dermaE (Table~\ref{tab:dermamnist}) show 
that it is indeed a more challenging dataset and is guaranteed to be void of any data leakage. 
\rev{It should be noted that the \dermaE dataset is almost the same as the official partitions of the ISIC 2018 Challenge data with 2 distinctions: the images in \dermaE are resized (\lowres or \hires) and the \say{easter egg} image has been removed from the testing partition.}

\ka{A summary of the three datasets: \dermamnist, \dermaC, and \dermaE is presented in Table~\ref{tab:datasets_summary}, listing a brief description and statistics of the datasets.}

\begin{figure}[!ht]
\centering
\includegraphics[width=\textwidth]{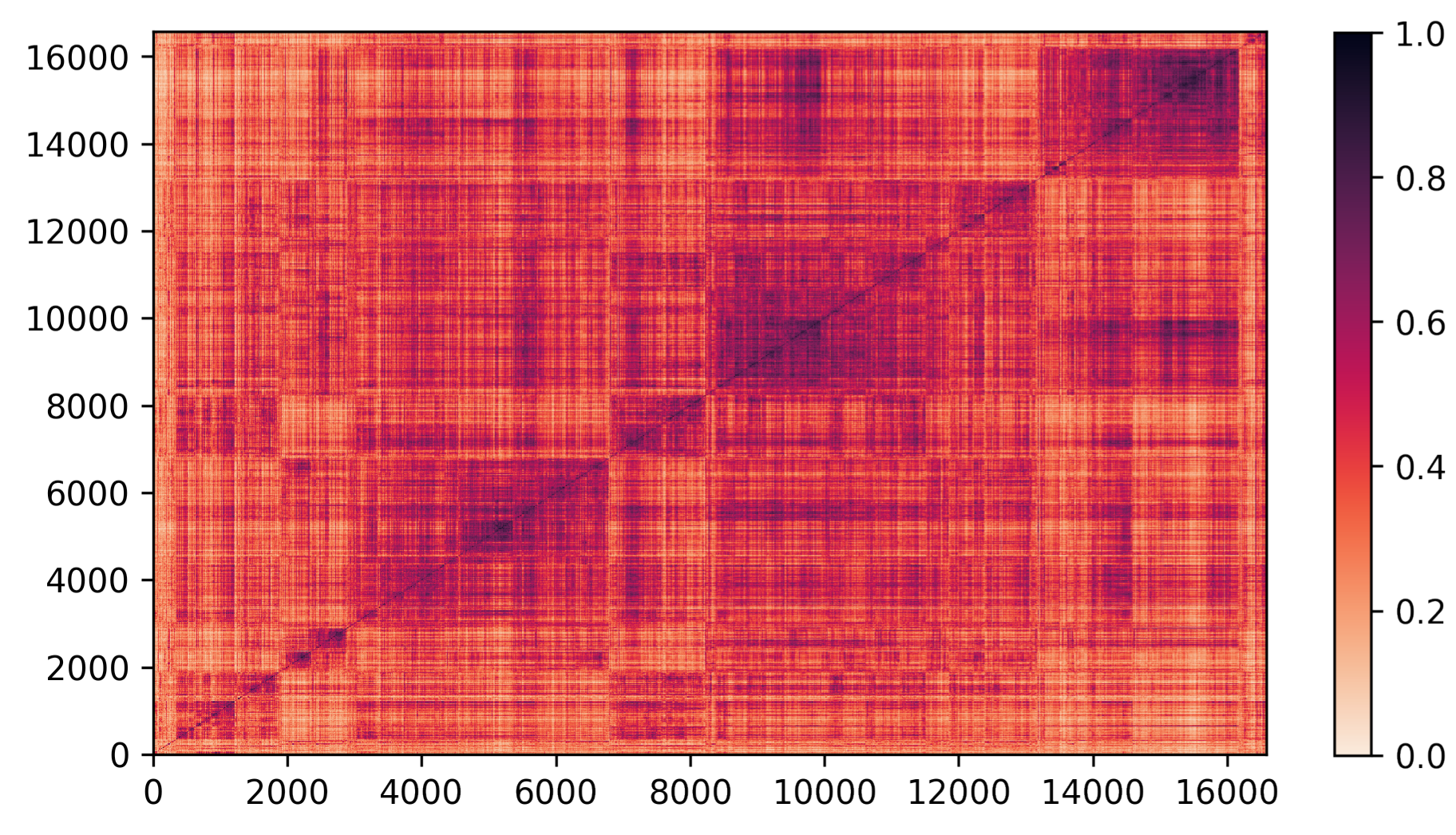}
\caption{Inter-image similarity matrix ($\fitzcount{} \times \fitzcount{}$) computed and visualized for all pairs of images in the \fitzdset dataset, where image pairs with higher similarity are represented by darker colors. Note how there are several regions of dark-colored pairs, indicating the presence of potential duplicates in the dataset.}
\label{fig:fitz_distmat_vis}
\end{figure}

\begin{figure*}[!ht]
     \centering
     \begin{subfigure}[t]{0.22\textwidth}
         \centering
         \includegraphics[width=\textwidth]{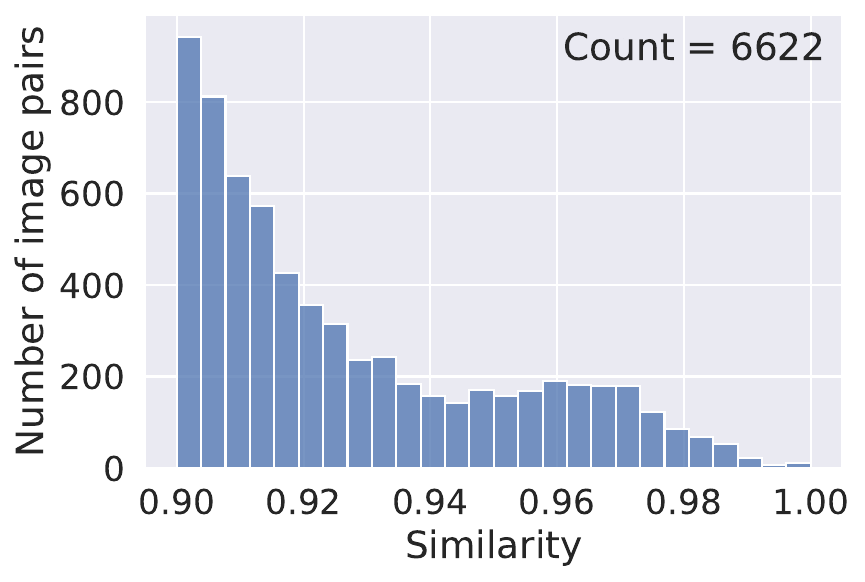}
         \caption{$\{\mathcal{S}_{0.90}\}$}
     \end{subfigure}
     \quad
     \begin{subfigure}[t]{0.22\textwidth}
         \centering
         \includegraphics[width=\textwidth]{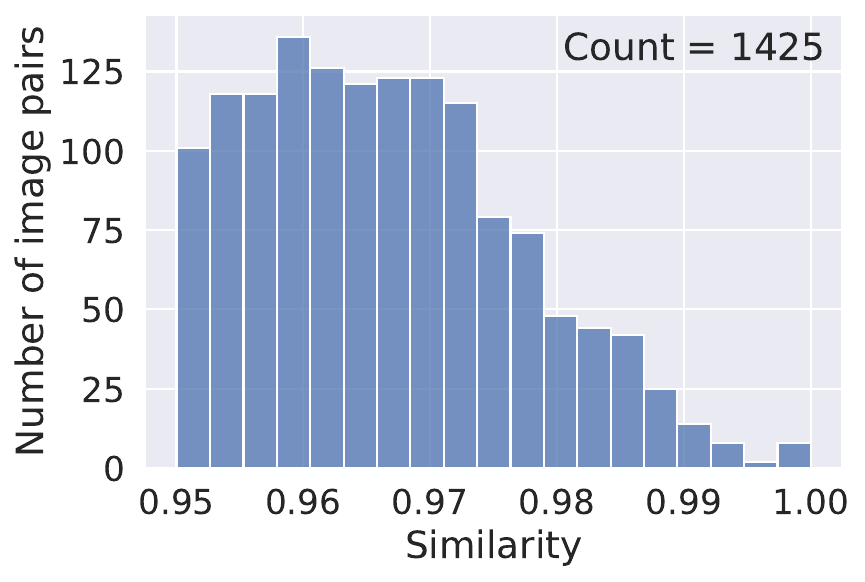}
         \caption{$\{\mathcal{S}_{0.95}\}$}
     \end{subfigure}
     \quad
     \begin{subfigure}[t]{0.22\textwidth}
         \centering
         \includegraphics[width=\textwidth]{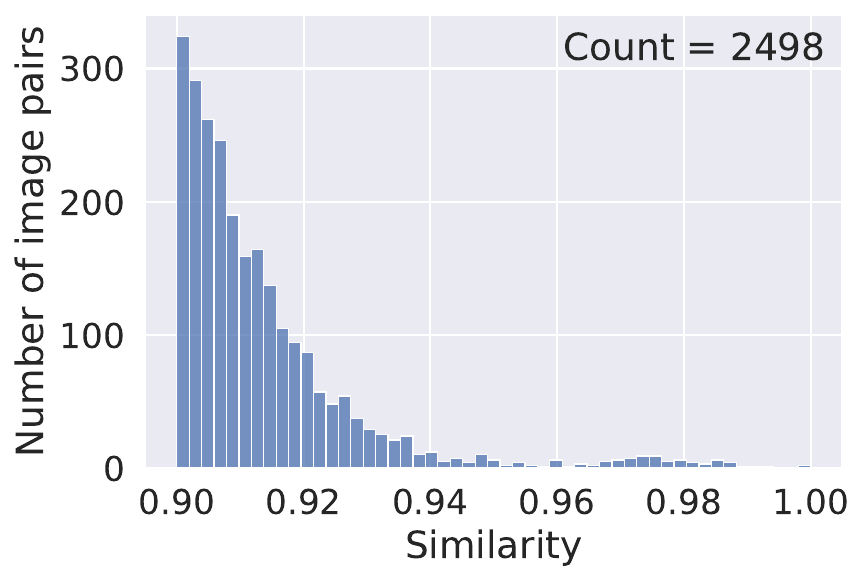}
         \caption{$\{\mathcal{S}_{0.90} \cap \hat{\mathcal{D}}\}$}
     \end{subfigure}
     \quad
     \begin{subfigure}[t]{0.22\textwidth}
         \centering
         \includegraphics[width=\textwidth]{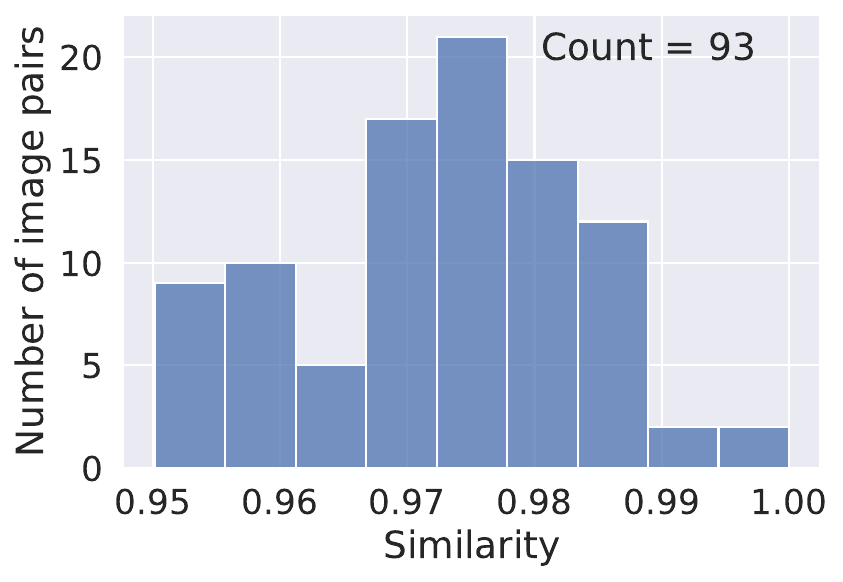}
         \caption{$\{\mathcal{S}_{0.95} \cap \hat{\mathcal{D}}\}$}
     \end{subfigure}
     \\
     \begin{subfigure}[t]{0.22\textwidth}
         \centering
         \includegraphics[width=\textwidth]{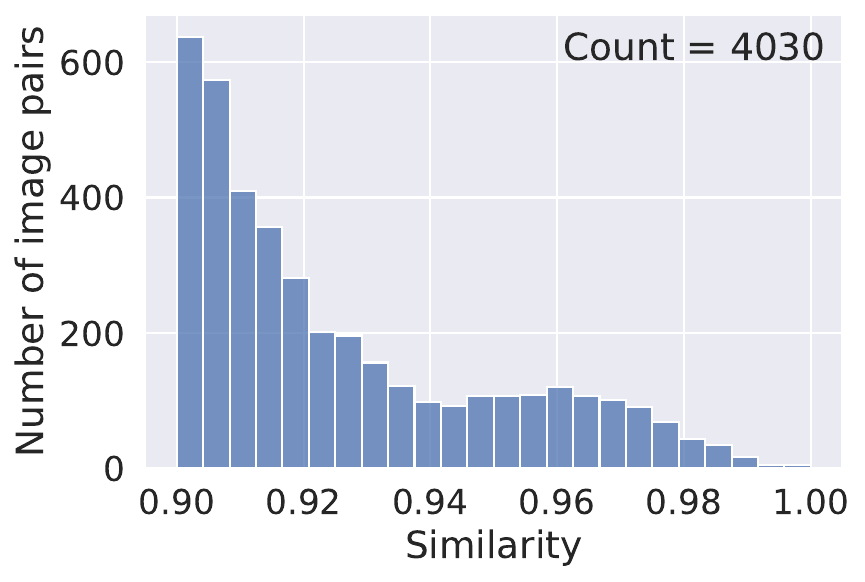}
         \caption{$\{\mathcal{S}_{0.90} \cap \diffF\}$}
     \end{subfigure}
     \quad
     \begin{subfigure}[t]{0.22\textwidth}
         \centering
         \includegraphics[width=\textwidth]{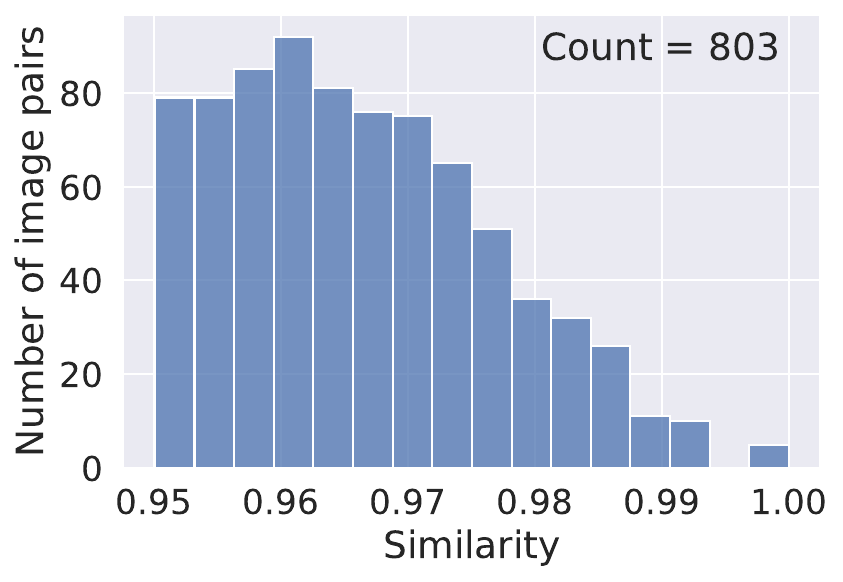}
         \caption{$\{\mathcal{S}_{0.95} \cap \diffF\}$}
     \end{subfigure}
     \quad
     \begin{subfigure}[t]{0.22\textwidth}
         \centering
         \includegraphics[width=\textwidth]{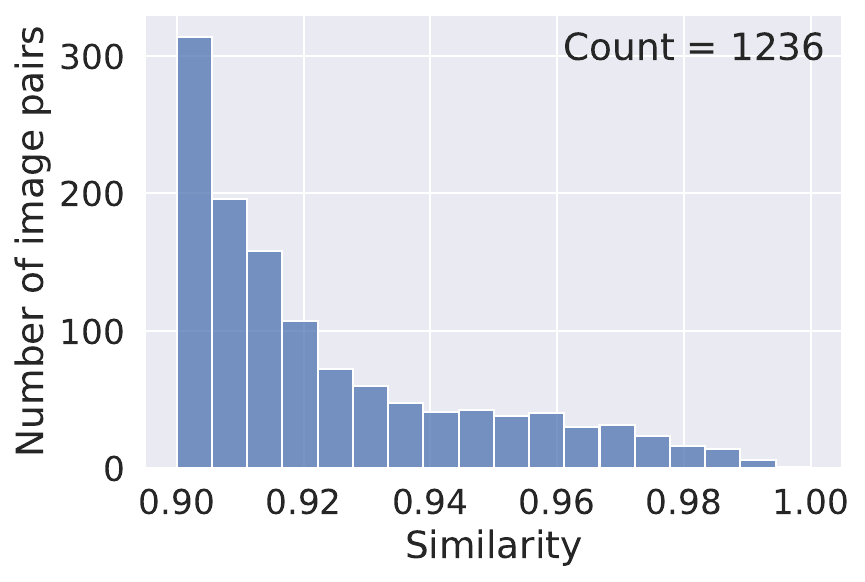}
         \caption{$\{\mathcal{S}_{0.90} \cap \diffFStrict\}$}
     \end{subfigure}
     \quad
     \begin{subfigure}[t]{0.22\textwidth}
         \centering
         \includegraphics[width=\textwidth]{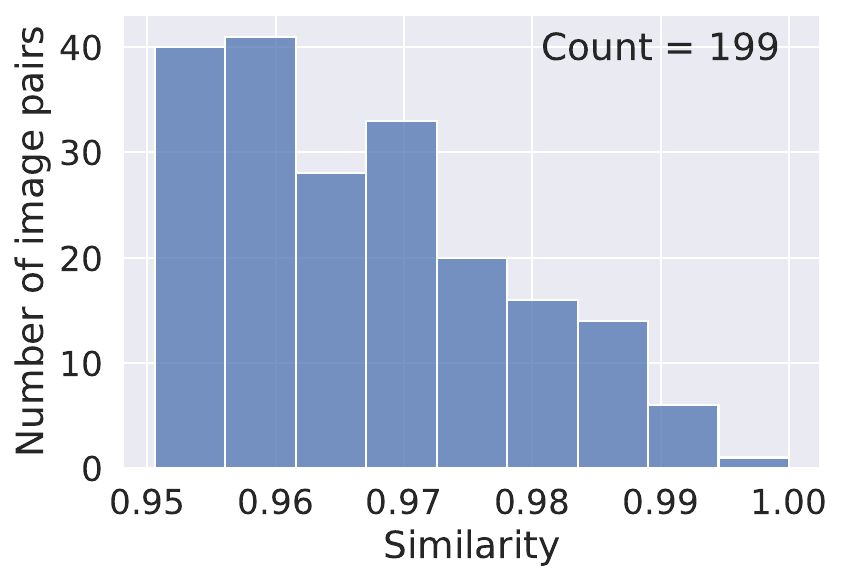}
         \caption{$\{\mathcal{S}_{0.95} \cap \diffFStrict\}$}
     \end{subfigure}
     \\
     \begin{subfigure}[t]{0.22\textwidth}
         \centering
         \includegraphics[width=\textwidth]{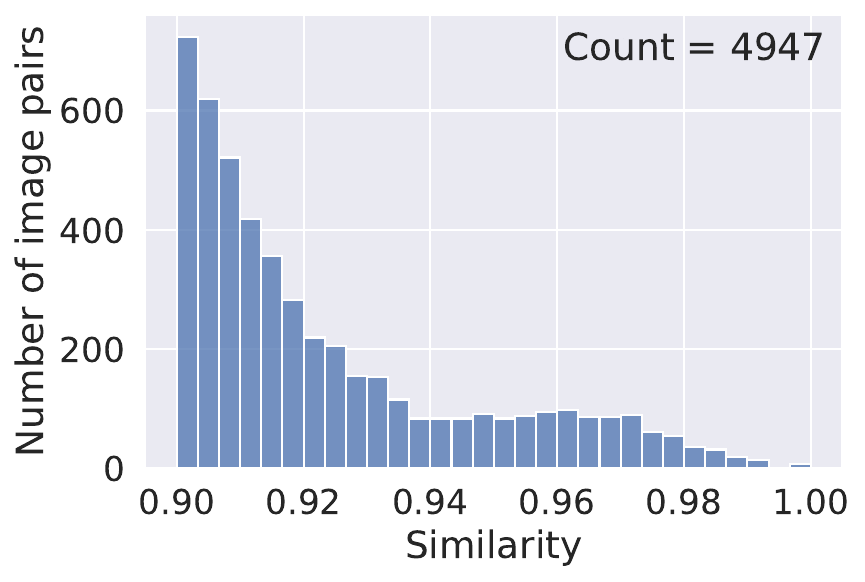}
         \caption{$\{\mathcal{S}_{0.90} \cap \{\hat{\mathcal{D}} \cup \diffF\}\}$}
     \end{subfigure}
     \quad
     \begin{subfigure}[t]{0.22\textwidth}
         \centering
         \includegraphics[width=\textwidth]{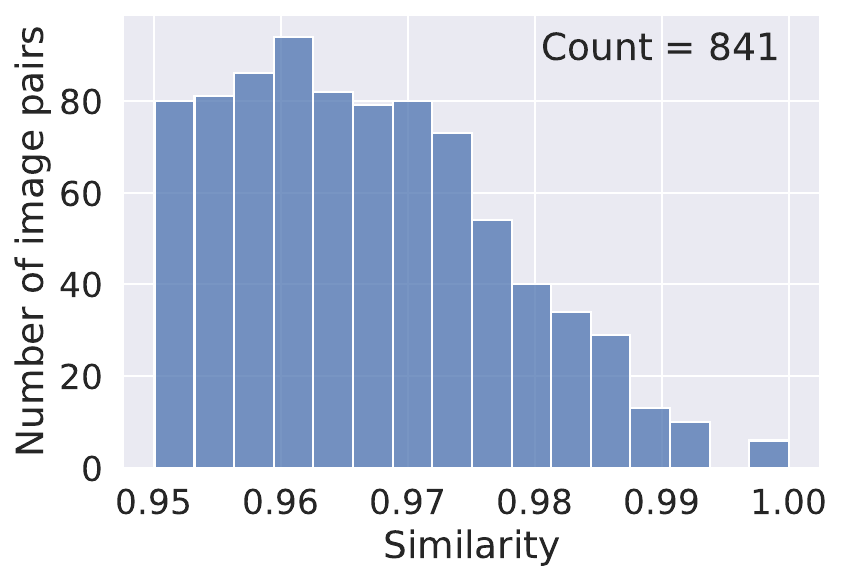}
         \caption{$\{\mathcal{S}_{0.95} \cap \{\hat{\mathcal{D}} \cup \diffF\}\}$}
     \end{subfigure}
     \quad
     \begin{subfigure}[t]{0.22\textwidth}
         \centering
         \includegraphics[width=\textwidth]{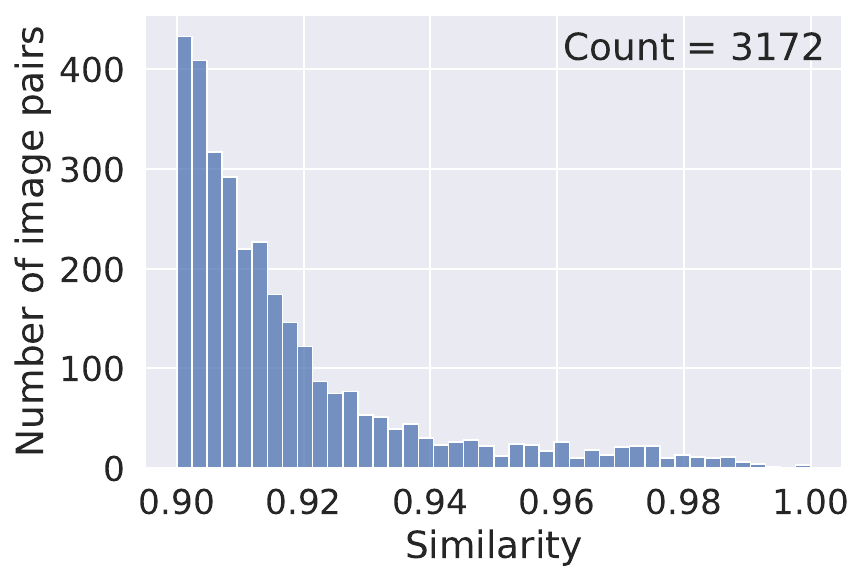}
         \caption{$\{\mathcal{S}_{0.90} \cap \{\hat{\mathcal{D}} \cup \diffFStrict\}\}$}
     \end{subfigure}
     \quad
     \begin{subfigure}[t]{0.22\textwidth}
         \centering
         \includegraphics[width=\textwidth]{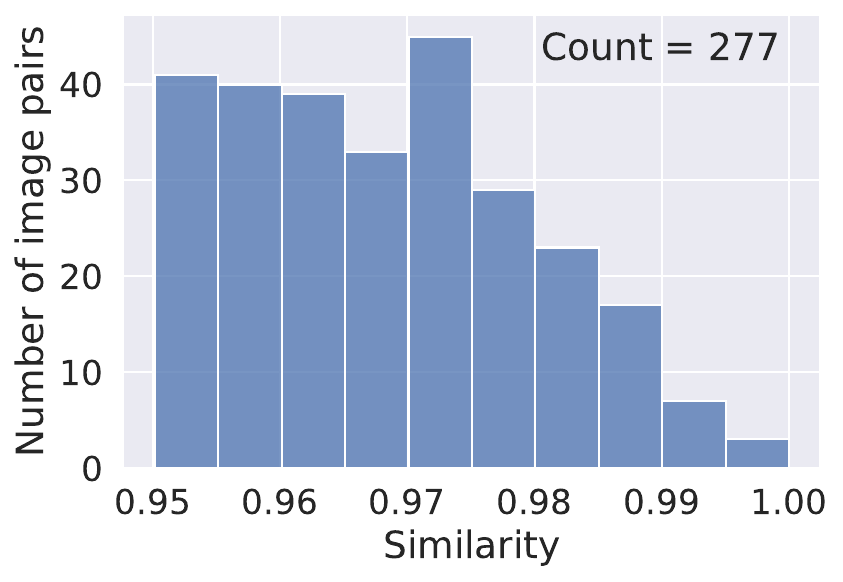}
         \caption{$\{\mathcal{S}_{0.95} \cap \{\hat{\mathcal{D}} \cup \diffFStrict\}\}$}
     \end{subfigure}
     \\
     \begin{subfigure}[t]{0.22\textwidth}
         \centering
         \includegraphics[width=\textwidth]{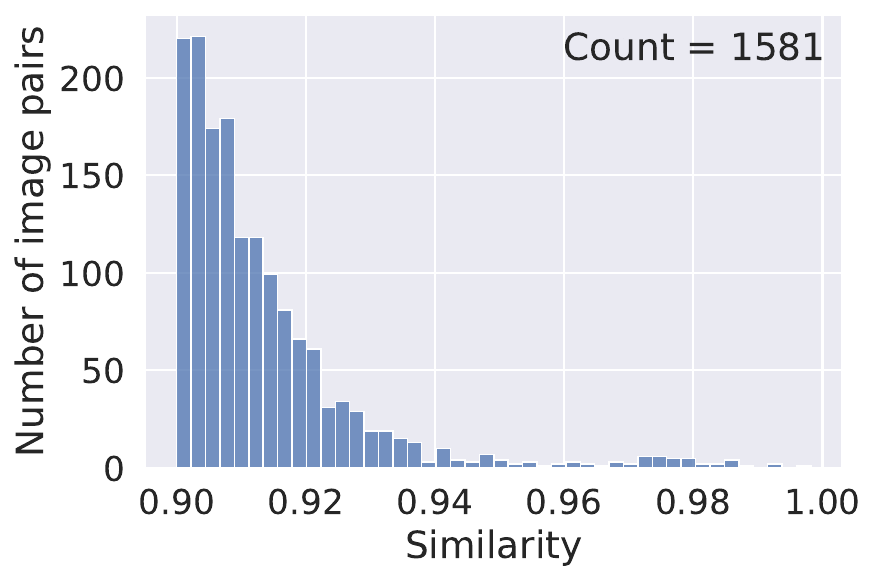}
         \caption{$\{\mathcal{S}_{0.90} \cap \{\hat{\mathcal{D}} \cap \diffF\}\}$}
     \end{subfigure}
     \quad
     \begin{subfigure}[t]{0.22\textwidth}
         \centering
         \includegraphics[width=\textwidth]{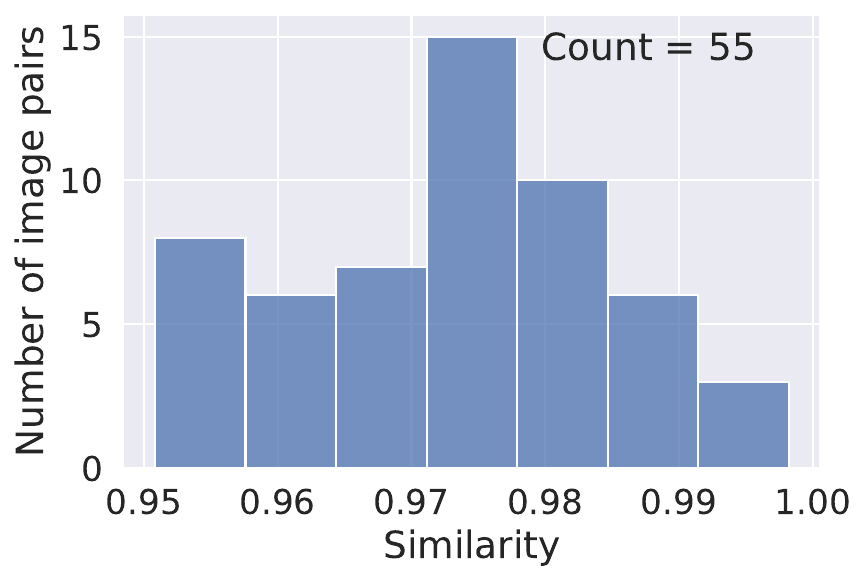}
         \caption{$\{\mathcal{S}_{0.95} \cap \{\hat{\mathcal{D}} \cap \diffF\}\}$}
     \end{subfigure}
     \quad
     \begin{subfigure}[t]{0.22\textwidth}
         \centering
         \includegraphics[width=\textwidth]{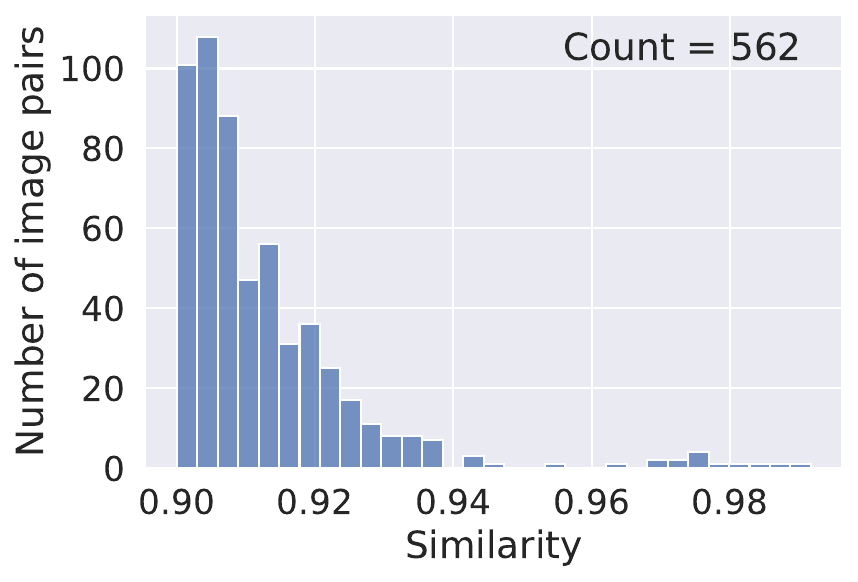}
         \caption{$\{\mathcal{S}_{0.90} \cap \{\hat{\mathcal{D}} \cap \diffFStrict\}\}$}
     \end{subfigure}
     \quad
     \begin{subfigure}[t]{0.22\textwidth}
         \centering
         \includegraphics[width=\textwidth]{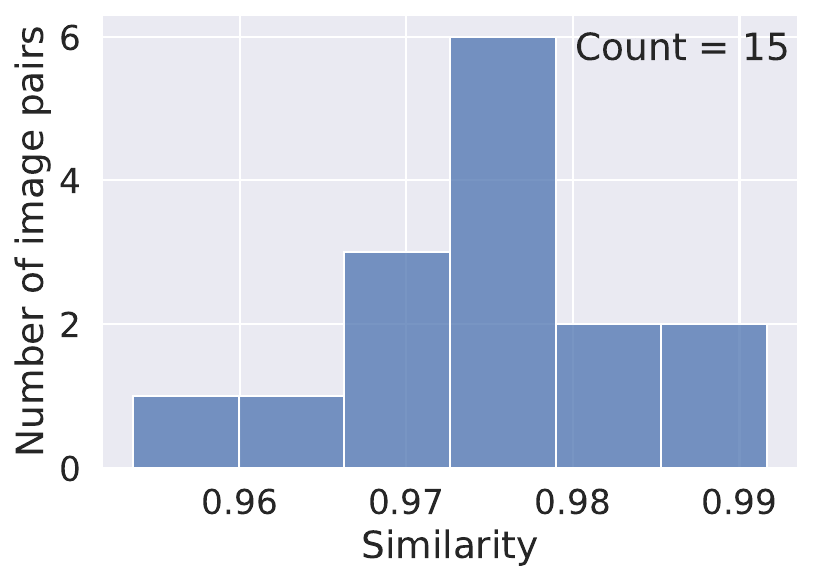}
         \caption{$\{\mathcal{S}_{0.95} \cap \{\hat{\mathcal{D}} \cap \diffFStrict\}\}$}
     \end{subfigure}

        \caption{
        Visualizing the distributions of duplicates in \fitzdset, filtered by different combinations of criteria. Total counts are inset in each plot.
        $\mathcal{S}_{0.90}$ and $\mathcal{S}_{0.95}$ denotes pairs with similarity scores of at least 0.90 and 0.95 respectively. $\hat{\mathcal{D}}$ denotes pairs that differ in their diagnoses labels. $\diffF$ and $\diffFStrict$ denotes pairs that `differ in their FST labels by 1' versus `by more than 1', respectively. Best viewed online.
        }
        \label{fig:fitzpatrick_stats}
\end{figure*}

\begin{figure}[!ht]
\centering
 \begin{subfigure}[b]{\textwidth}
         \centering
         \includegraphics[width=\textwidth]{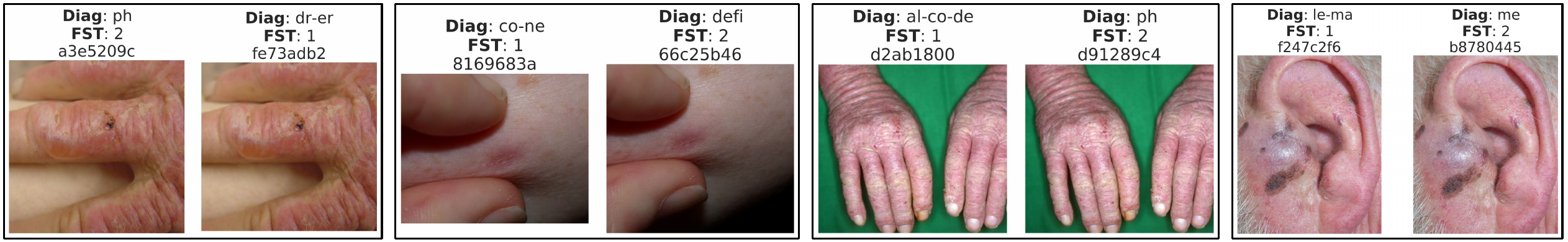}
         \caption{Duplicate image pairs with a very high degree of similarity ($\mathcal{S} > 0.95$) that differ in both their diagnosis as well as FST labels ($\{\mathcal{S}_{0.95} \cap \{\hat{\mathcal{D}} \cap \diffF\}\}$).}
     \end{subfigure}
     \\
     \vspace{2mm}
     \begin{subfigure}[b]{\textwidth}
         \centering
         \includegraphics[width=\textwidth]{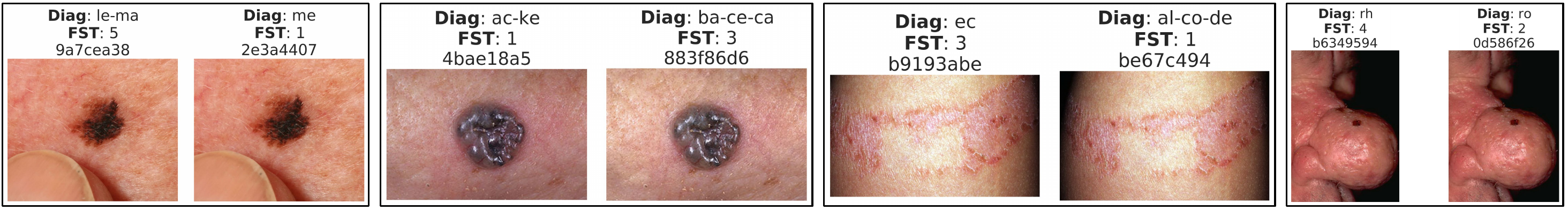}
         \caption{Duplicate image pairs with a very high degree of similarity ($\mathcal{S} > 0.95$) that have different diagnosis and strictly different FST labels $\{\mathcal{S}_{0.95} \cap \{\hat{\mathcal{D}} \cap \diffFStrict\}\}$.}
     \end{subfigure}
     \\
     \vspace{2mm}
     \begin{subfigure}[b]{\textwidth}
         \centering
         \includegraphics[width=\textwidth]{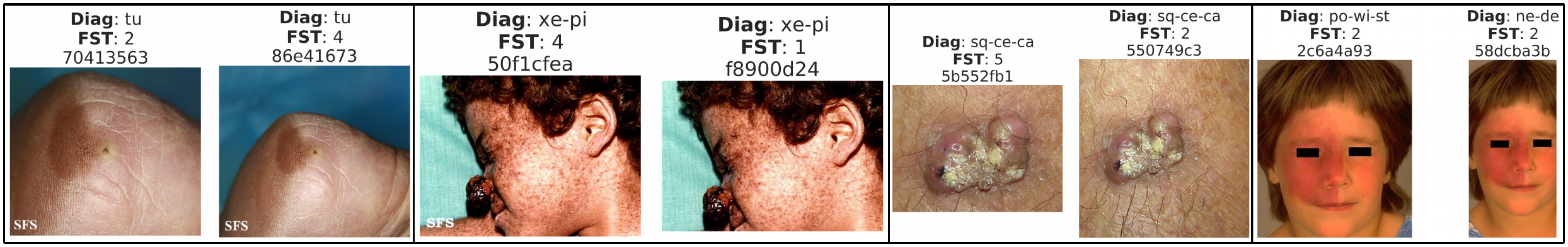}
         \caption{Duplicate image pairs, where one image is a zoomed in and/or a cropped version of the other.}
     \end{subfigure}
     \\
     \vspace{2mm}
     \begin{subfigure}[b]{\textwidth}
         \centering
         \includegraphics[width=\textwidth]{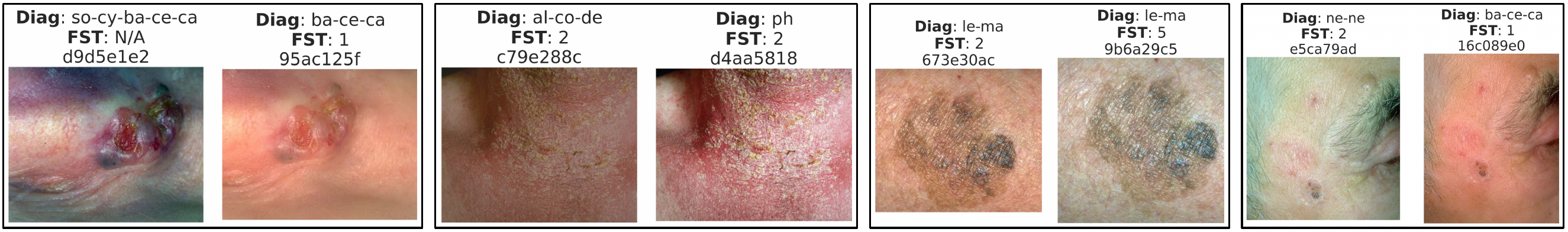}
         \caption{Duplicate image pairs, where the illumination settings are different for the images (e.g., with and without the camera flash).}
     \end{subfigure}
     \\
     \vspace{2mm}
     \begin{subfigure}[b]{\textwidth}
         \centering
         \includegraphics[width=\textwidth]{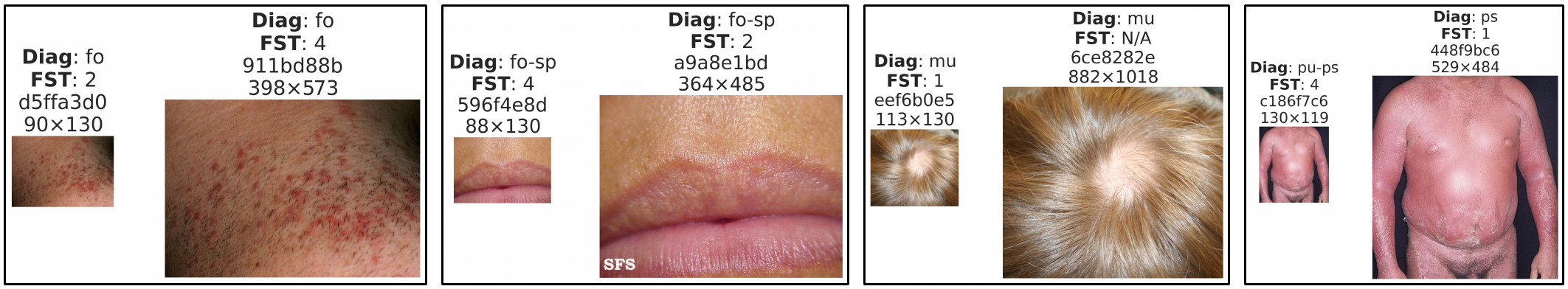}
         \caption{Duplicate image pairs, where the first image is a low-resolution copy of the second image. The resolutions of both images in all pairs is denoted, and image sizes displayed here are not representative of the relative spatial resolutions.}
     \end{subfigure}
     
     \caption{ 
     Samples of duplicate image pairs from \fitzdset, with their corresponding diagnosis (Diag.) and FST labels. For duplicate pairs where one image is a lower-resolution of the other \textbf{(e)}, the spatial resolutions of the images have also been displayed. Notice several image pairs, including all the images in \textbf{(a, b)} are exact copies. More concerningly, these duplicate images can vary in their diagnosis and/or FST labels, the latter by as much as 4 skin tones. Best viewed online. \textbf{(Continued in Fig.~\ref{fig:fitz_duplicates_part2})}
     }
    \label{fig:fitz_duplicates}
     
     \end{figure}

\begin{figure}[!ht]
\centering
 \begin{subfigure}[b]{\textwidth}
         \centering
         \includegraphics[width=\textwidth]{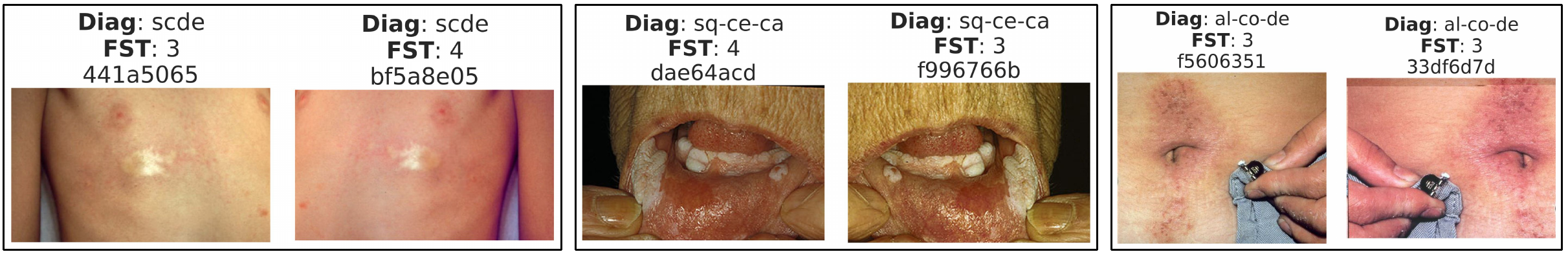}
         \caption{Duplicate image pairs, where a simple geometrical transformation (e.g., mirroring or lateral inversion) has been applied to one image to obtain the other.}
     \end{subfigure}
     \\
     \vspace{2mm}
     \begin{subfigure}[b]{\textwidth}
         \centering
         \includegraphics[width=\textwidth]{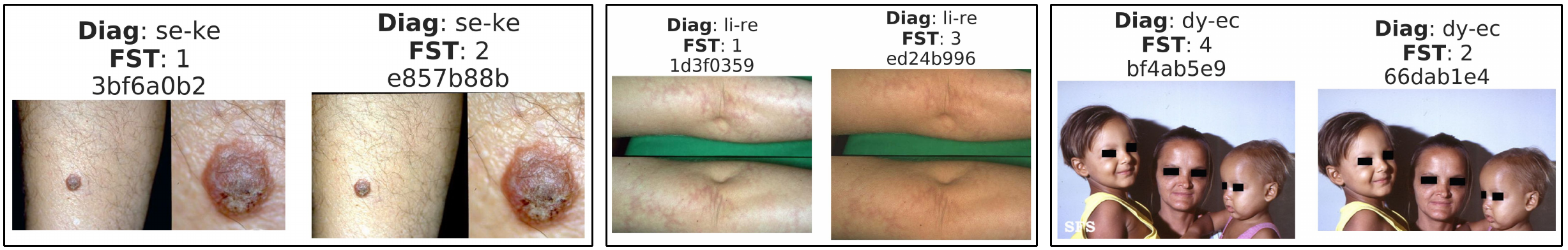}
         \caption{Duplicate image pairs with more than one object of interest or multiple people. With the latter, it is difficult to ascertain which individual's diagnosis and FST has been labeled.}
     \end{subfigure}
     \\
     \vspace{2mm}
     \begin{subfigure}[b]{\textwidth}
         \centering
         \includegraphics[width=\textwidth]{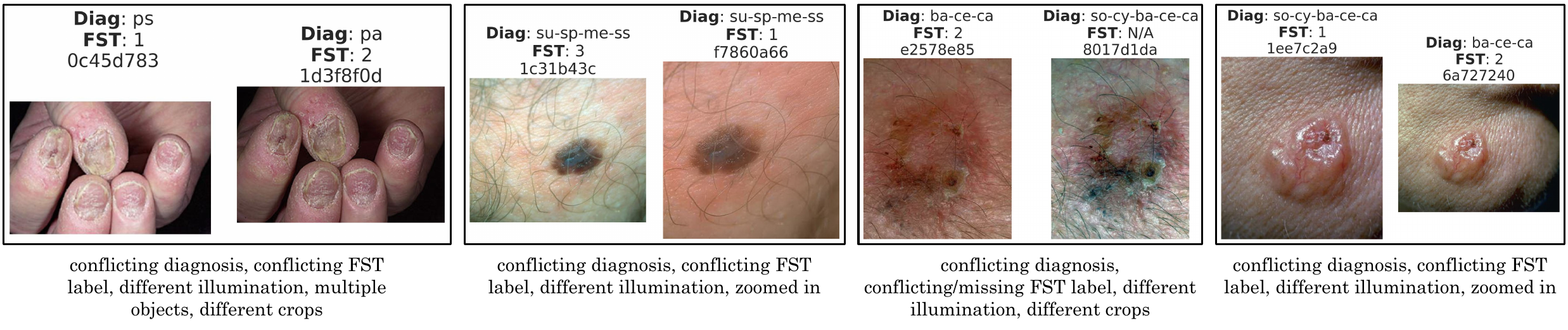}
         \caption{Duplicate image pairs exhibiting more than one of the aforementioned issues.}
     \end{subfigure}
     \\
     \vspace{2mm}
     \begin{subfigure}[b]{\textwidth}
         \centering
         \includegraphics[width=\textwidth]{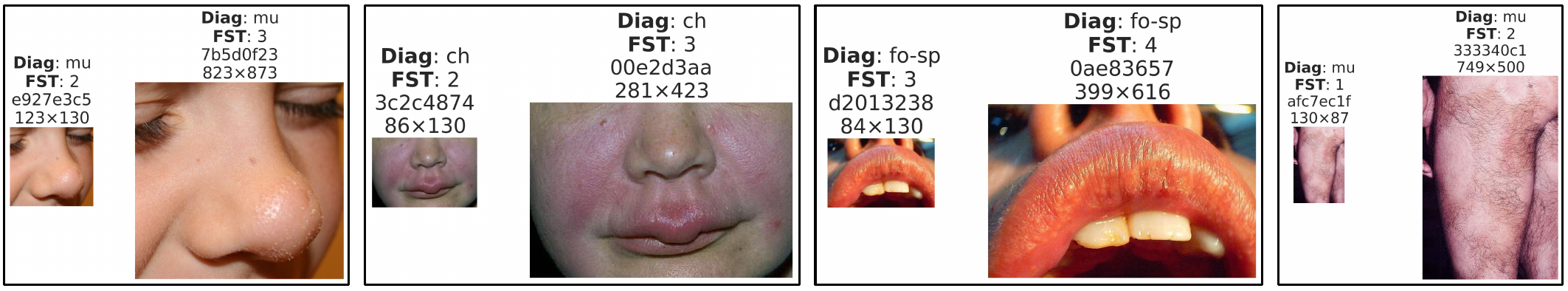}
         \caption{Duplicates detected by \pypackage{cleanvision}. All the 19 duplicate pairs detected had one image as a lower-resolution copy of the other, similar to the pairs in \textbf{(e)}.}
     \end{subfigure}
     
     \caption{ 
     \textbf{(Continued from Fig.~\ref{fig:fitz_duplicates})}
     Samples of duplicate image pairs from \fitzdset, with their corresponding diagnosis (Diag.) and FST labels. For duplicate pairs where one image is a lower-resolution of the other \textbf{(d)}, the spatial resolutions of the images have also been displayed. Best viewed online.
     }
     
    \label{fig:fitz_duplicates_part2}
     
     \end{figure}

\begin{figure}[!ht]
\centering
\includegraphics[width=\textwidth]{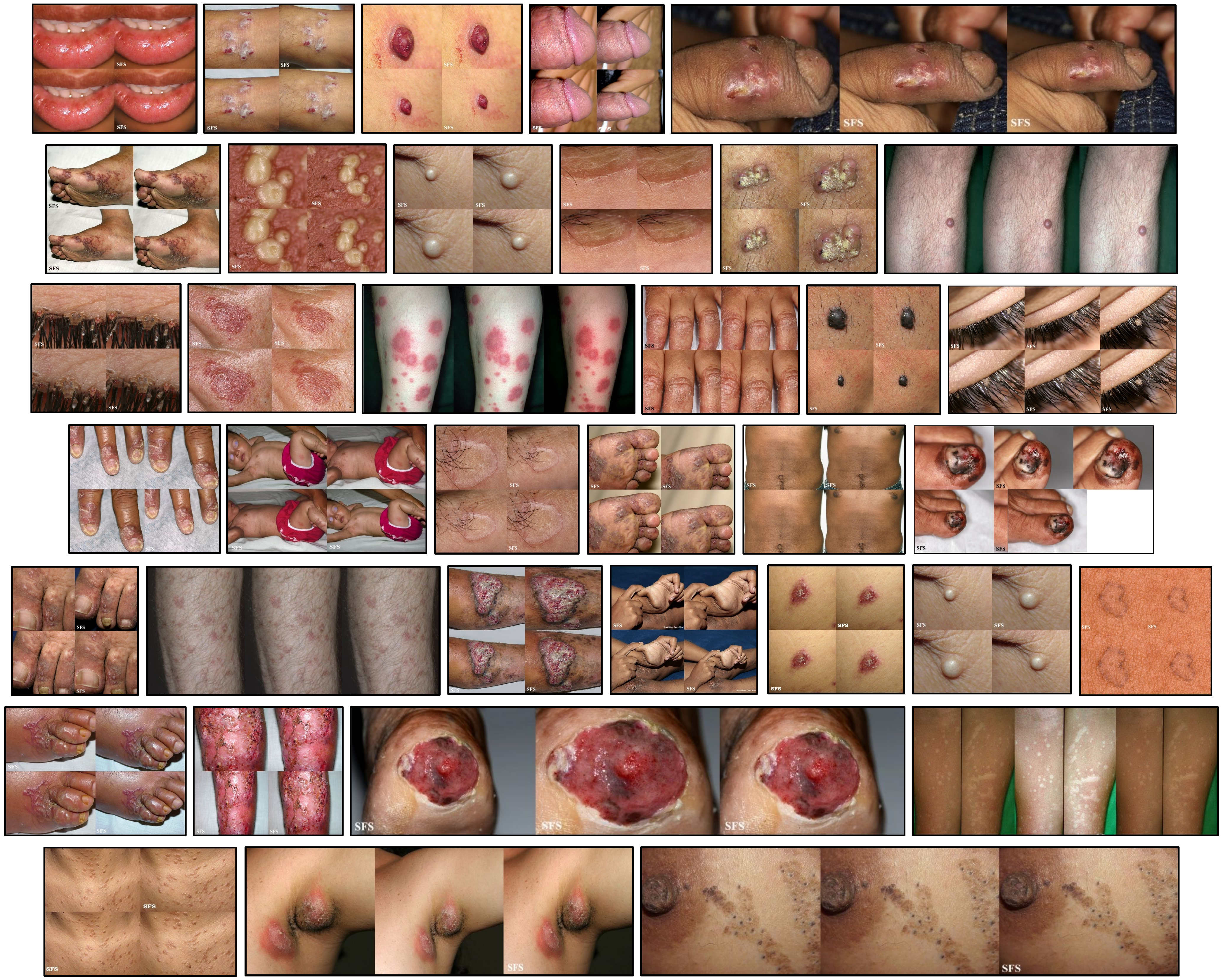}
\caption{Sample clusters 
of duplicate images from \fitzdset with intra-cluster similarity $\ge$ 0.90. Just like Fig.~\ref{fig:fitz_duplicates}, these images are either near or exact copies, with slight differences originating from different crops, altered aspect ratios, camera flash, and small changes to the viewpoint. Best viewed online.}
\label{fig:fitz_cc}
\end{figure}

\begin{figure}[!ht]
\centering
\includegraphics[width=0.85\textwidth]{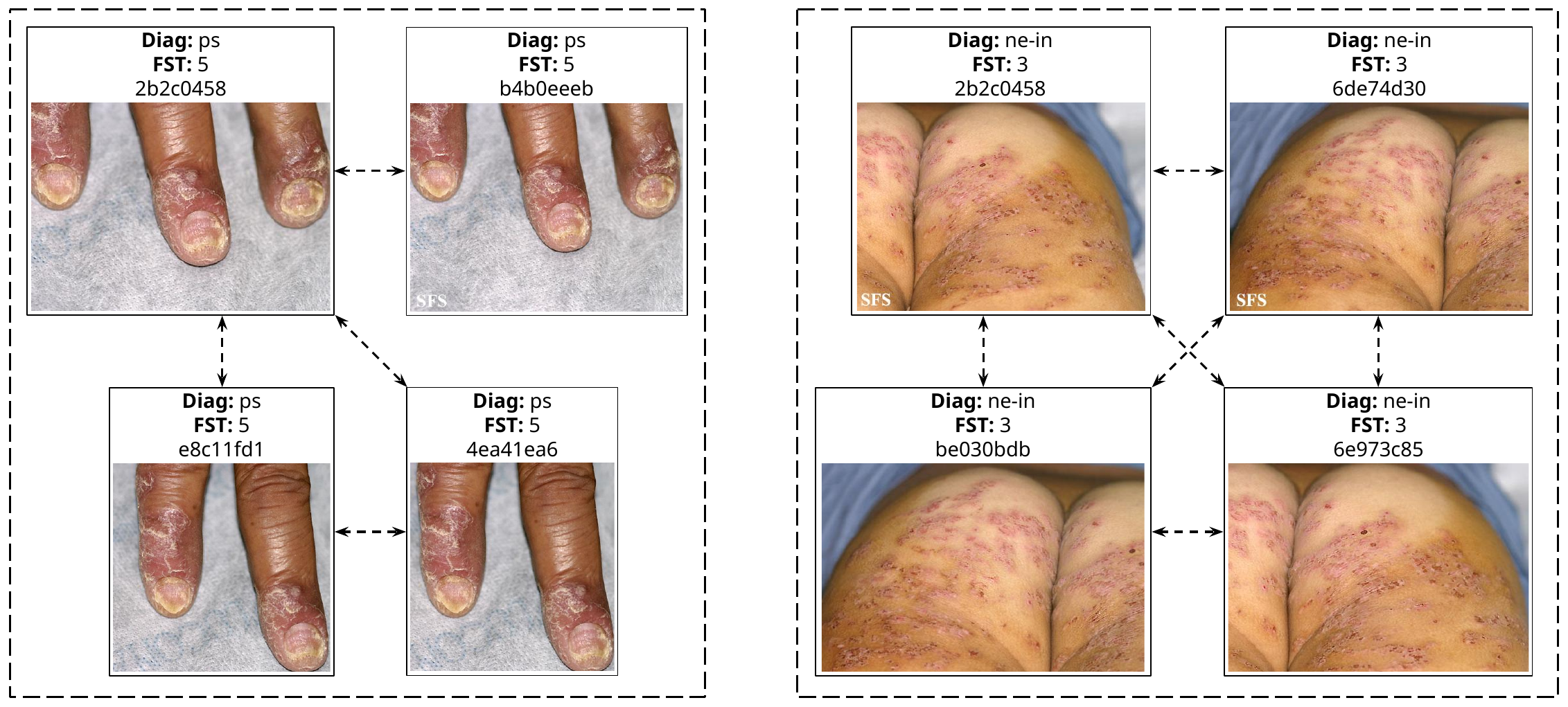}
\caption{
\rev{Merging duplicate image pairs to form larger clusters of duplicate images in \fitzdset. The arrows show pairs of duplicates detected: there are 4 duplicate pairs on the left and 6 duplicate pairs on the right. Merging them leads to 2 clusters of duplicates images, with 4 images in each cluster.}
}
\label{fig:f17k_coalesce_duplicates}
\end{figure}

\begin{figure}[!ht]
     \centering

     \begin{subfigure}[t]{0.65\textwidth}
         \centering
         \includegraphics[width=\textwidth]{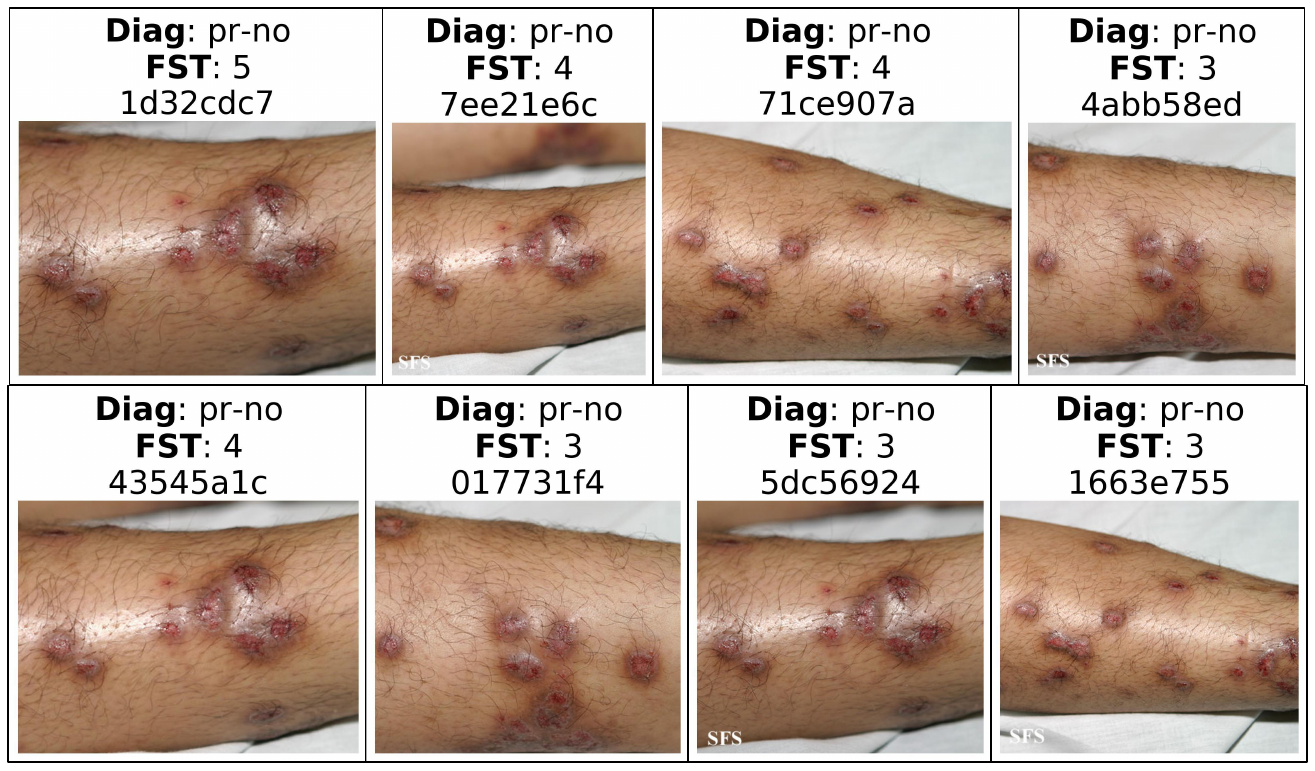}
         \phantomsubcaption
     \end{subfigure}
     \\
     \vspace{4mm}
     \begin{subfigure}[t]{0.8\textwidth}
         \centering
         \includegraphics[width=\textwidth]{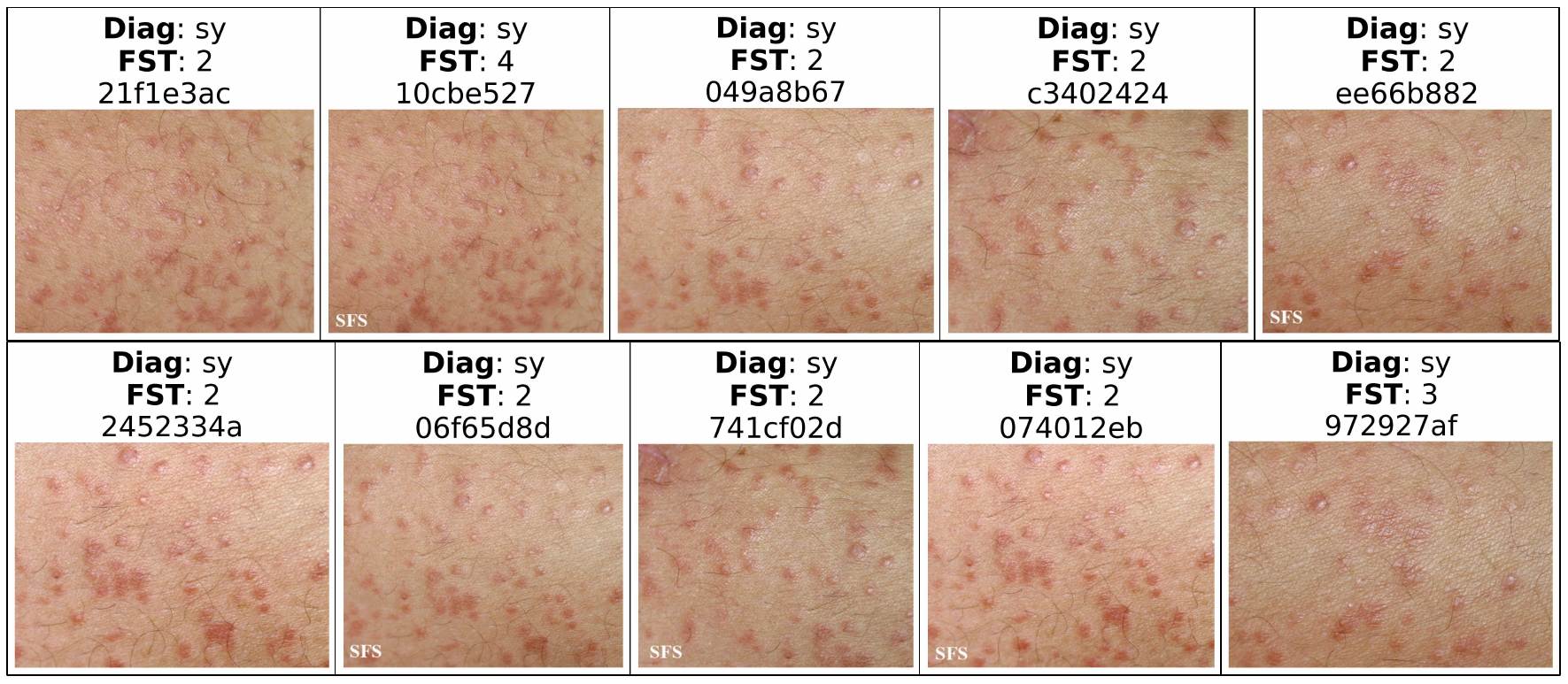}
         \phantomsubcaption
     \end{subfigure}

    \caption{
    Examples of large clusters of duplicates detected in \fitzdset, containing as many as 10 duplicate images.
    }
    \label{fig:f17k_large_cc}
\end{figure}

\subsection{\fitzdset}
\label{subsec:fitz_analysis}

Released in 2021, \fitzdset is one of the largest publicly available datasets of clinical skin disease images. The large number of skin diseases covered (114), the in-the-wild nature of the images therein, and the availability of associated and diverse \fitz skin tone (FST) labels~\cite{alipour2023skin}
make it an immensely valuable dataset for skin image analysis research.
However, unlike 
\dermamnist, 
which is collected from clinical visits and whose labels were confirmed, \fitzdset was curated from 2 publicly available online dermatological atlases: DermaAmin~\cite{dermaamin} (12,672 images) and Atlas Dermatologico~\cite{atlasdermatologico} (3,905 images).
As such, the diagnosis labels of these images are not confirmed, through histopathology or otherwise. 
The authors conducted a small-scale study on only 3.04\% of the entire dataset (504 out of \fitzcount{} images) where 2 board-certified dermatologists assessed the diagnoses of the images, and the consensus was that only 69.0\% of the images were clearly diagnostic of the disease label and, more importantly, 3.4\% of the images were mislabeled. 
This is problematic since \fitzdset has been used to train models for high\revnewer{-}stakes applications such as model explainability~\cite{abid2022meaningfully,ghadiri2024xtranprune}, trustworthiness~\cite{yan2023trustable}, skin tone detection~\cite{bevan2022detecting},
\revnewer{model calibration~\cite{yuksekgonul2023beyond}}
and fairness~\cite{du2022fairdisco,pakzad2022circle,seth2024does,wang2024achieving,chiu2024achieve,aayushman2024fair}.
\revnewer{\fitzdset has also been used for for training and evaluating large vision-language models~\cite{kim2024transparent,akuffo2024assessing,chen2024gmai,yan2024general,zhou2024pretrained},
visual question answering~\cite{hu2024omnimedvqa,yim2024dermavqa}, 
clinical decision support for differential diagnosis~\cite{groh2024deep},
generative modeling~\cite{sagers2022improving,lin2024skingen,mekala2024synthetic},
federated learning~\cite{alhamoud2024fedmedicl},
and for creating a derivative dataset: SkinCon~\cite{daneshjou2022skincon}}. 
Pakzad et al.~\cite{pakzad2022circle} previously highlighted the existence of erroneous and wrongly labeled images in \fitzdset, and
for these reasons, we investigate the extent of labeling inaccuracy in this dataset.

\subsubsection{Data duplication and leakage}

To investigate the presence of duplicates in \fitzdset,
we use \pypackage{fastdup}
to 
calculate inter-image embedding similarity scores $\mathcal{S} (x_i, x_j)$ for all
$\binom{\fitzcount{}}{2}$
pairs of images \rev{$(x_i, x_j)$}. These are shown as a $\fitzcount{} \times \fitzcount{}$ similarity matrix in Fig.~\ref{fig:fitz_distmat_vis}, and we can clearly see that there are several pairs with high similarities, denoted by darker shades, spread throughout the dataset.
For subsequent analyses, we restrict ourselves to pairs with a high similarity by setting thresholds to 
$\mathcal{S} \ge \tau; \ \tau \in \{0.90, 0.95\}$.
The distributions of image embedding pairs with these similarity thresholds are shown in Fig.~\ref{fig:fitzpatrick_stats} (a, b), respectively. Note that there are 6,622 and 1,425 image embedding pairs with similarity scores greater than 0.90 and 0.95, respectively. Manual verification by a human reviewer of these
1,425 image pairs, whose embeddings had similarities greater than 0.95, 
revealed that 98.39\% of these images (1,402 images) were indeed duplicates (Fig.~\ref{fig:fitz_duplicates}, Fig.~\ref{fig:fitz_duplicates_part2}), with 16 pairs (1.12\%) being false positive and 7 pairs (0.49\%) being ambiguous. A second reviewer agreed with 1,419 labels of the first reviewer (99.58\% match), exhibiting a near-perfect agreement~\cite{cohen1960coefficient,landis1977measurement} (Cohen's kappa $\kappa = 0.87$).
We visualize some of these pairs in Fig.~\ref{fig:fitz_duplicates} and Fig.~\ref{fig:fitz_duplicates_part2}, categorizing them according to traits exhibited by the pairs. Since the filenames of the images in \fitzdset are of the format \texttt{\texttt{\textlangle MD5hash\textrangle.jpg}}, for each image, we also display their diagnosis abbreviation, their FST label (set to `N/A' when the FST label is missing), and a truncated MD5 hash to uniquely identify the images. 
Notice that duplicate image pairs exist because of:
\begin{itemize}
    \item different crop/zoom levels (Fig.~\ref{fig:fitz_duplicates} (c)),
    \item different illumination setups (Fig.~\ref{fig:fitz_duplicates} (d)),
    \item different image resolutions (Fig.~\ref{fig:fitz_duplicates} (e)), and 
    \item simple geometrical transformations (e.g., mirroring) (Fig.~\ref{fig:fitz_duplicates_part2} (a)).
\end{itemize}
Worryingly, duplicate image pairs containing multiple disjoint objects of interest or multiple people also exist (Fig.~\ref{fig:fitz_duplicates_part2} (b)), making it difficult to determine to which of these the diagnosis and the FST labels apply. Finally, several duplicate pairs with more than one of these issues were also detected (Fig.~\ref{fig:fitz_duplicates_part2} (c)).

For a more 
\rev{detailed}
duplicate detection, we employ another Python library, \pypackage{cleanvision}~\cite{cleanvision_cite},
to further assess the dataset. Aside from the duplicate image pairs found by \pypackage{fastdup} based on our similarity threshold of $\mathcal{S} \ge 0.90$, the \pypackage{cleanvision} analysis found 19 more duplicate pairs that have slightly lower inter-image similarity ($0.85 \le \mathcal{S} < 0.90$), primarily because of the large difference in spatial resolutions between duplicate pairs (Fig.~\ref{fig:fitz_duplicates_part2} (d)).

Unfortunately, data duplication in \fitzdset is not limited to image pairs only.
We use \pypackage{fastdup} to cluster images whose intra-cluster image similarity is greater than 0.90\rev{, i.e., clusters of images where the mean similarity for all the image pairs is greater than 0.90}. We consider clusters of at least 3 images\rev{,}
since those with 2 images \rev{(i.e., duplicate images pairs)} are already covered in our analysis of duplicate pairs. \rev{Mathematically, we find all image clusters $\{x_1, x_2, \cdots, x_N\}; \ N \ge 3$ such that $\frac{1}{{\binom{N}{2}}} \sum_{i=1}^{N} \sum_{j=i+1}^{N} \mathcal{S}(x_i, x_j) > 0.90$}.
Manual verification of the clustering outputs yielded 139 image clusters, with $3.71 \pm 1.11$ images per cluster on average. 
Visual inspection of these image clusters (Fig.~\ref{fig:fitz_cc}), shows that these image clusters exhibit similar traits as the duplicate pairs, i.e., the images in each cluster are one or more of: exact matches, zoomed-in or cropped-out duplicates, duplicates with different illumination setups (captured with and without camera flash), or acquired at slightly different viewing angles. Finally, we merge the results of duplicate pairs with duplicate clusters, forming larger clusters as they are discovered. \rev{Fig.~\ref{fig:f17k_coalesce_duplicates} shows two examples of how several duplicate pairs are merged to form larger clusters of duplicate images.} This results in some large duplicate image clusters with as many as 10 images in a cluster (Fig.~\ref{fig:f17k_large_cc}), and a total of 2,297 clusters with $2.18 \pm 0.66$ images per cluster on average.

\subsubsection{Mislabeled diagnosis and FST labels}

\begin{figure}[!ht]
\centering
\includegraphics[width=\textwidth]{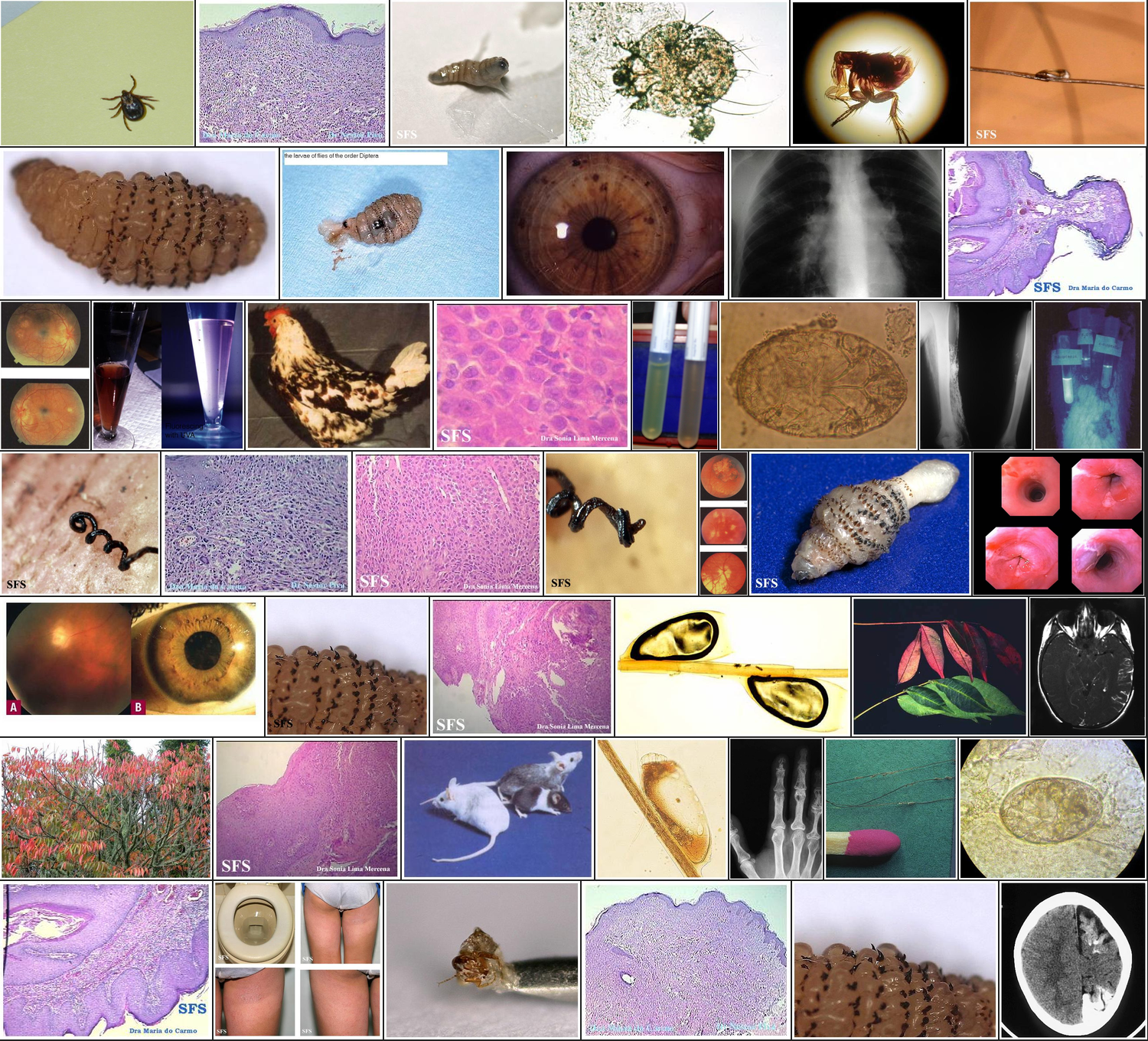}
\caption{Sample erroneous images or outliers in \fitzdset. Observe how the dataset contains several non-dermatological images, \eg images of plants, animals, and other medical imaging modalities.}
\label{fig:fitz_outliers}
\end{figure}

In addition to the presence of duplicates, \fitzdset contains images with mislabeled diagnoses and FSTs. 
Fig.~\ref{fig:fitz_duplicates} and Fig.~\ref{fig:fitz_duplicates_part2} show samples that are 
nearly identical
but differ in their diagnoses ($\mathcal{D}$) as well as FST ($\mathcal{F}$) labels, with the latter varying as much as 4 skin tones (FST 5 versus FST 1; Fig.~\ref{fig:fitz_duplicates} (b)).
For a more concrete estimation of the extent of mislabeling, we
use similarity thresholds of 0.90 (denoted by $\mathcal{S}_{0.90}$) and 0.95 (denoted by $\mathcal{S}_{0.95}$). We report the number of image pairs that have a similarity higher than \{0.90, 0.95\} but differ in their labels. Further, given the subjectivity of FST labels, Groh et al.~\cite{groh2021evaluating} evaluated the accuracy of human annotations (HA) to the gold standard (GT) subset using two metrics: accuracy and \say{off-by-one} accuracy, where the latter considers an annotation to be correct if 
{$|\mathcal{F}_{\mathrm{HA}} - \mathcal{F}_{\mathrm{GT}}| \leq 1$}.
Similar to previous works
that accounted for this \say{off-by-one} margin~\cite{bevan2022detecting,pakzad2022circle}, 
we count similar images that differ 
by at least 1 
($\diffF$)
and those that differ by strictly more than 1
FST
($\diffFStrict$).
Figs.~\ref{fig:fitz_duplicates} (a) and (b) show sample pairs with very high inter-image similarity ($\mathcal{S} > 0.95$) that differ in diagnoses and FST by 1 and more than 1, respectively.

Fig.~\ref{fig:fitzpatrick_stats} shows the distributions of duplicate image pairs filtered by one or more of: their similarity scores, whether their diagnoses are different, and whether and how much their FST scores differ by.
For image pair similarity thresholds of [0.90; 0.95], there are [2498; 93] image pairs that differ in their diagnoses 
($\hat{\mathcal{D}}$).
[4030; 803] image pairs differ in their FST labels ($\diffF$), while [1236; 199] pairs strictly differ in their FST labels ($\diffFStrict$). 
[4947; 841] images pairs differ in either their diagnosis or in their FST label ($\{\hat{\mathcal{D}} \cup \diffF\}$), and [3172; 277] differ in their diagnosis or strictly differ in the FST label ($\{\hat{\mathcal{D}} \cup \diffFStrict\}$). Finally, there are [1581; 55] and [562; 15] image pairs for the $\{\hat{\mathcal{D}} \cap \diffF\}$ and $\{\hat{\mathcal{D}} \cap \diffFStrict\}$ categories, respectively.

\begin{table}[!ht]
\centering
\resizebox{\textwidth}{!}{%
\setlength{\tabcolsep}{0.7em}
\def\arraystretch{1.25}
\begin{tabular}{lccccccc}
\toprule
\textbf{Holdout Set} & Verified & \begin{tabular}[c]{@{}c@{}}Random\\(Stratified)\end{tabular} & \begin{tabular}[c]{@{}c@{}}Source A\\(Atl. Derm.)\end{tabular} & \begin{tabular}[c]{@{}c@{}}Source B\\(DermaAmin)\end{tabular} & FST 3--6 & FST 1--2 \& 5--6 & FST 1--4 \\
\hline \# Train Images & 10,060 & 7,975 & 6,129 & 2,175 & 5,340 & 3,312 & 935 \\
\# Validation Images & 1,119 & 1,139 & 703 & 220 & 599 & 337 & 122 \\
\# Test Images & 215 & 2,280 & 2,395 & 6,832 & 5,029 & 6,685 & 6,903 \\
\midrule

\begin{tabular}[l]{@{}l@{}}Best\\Hyperparameters\\(\texttt{n\_epochs}, \texttt{optim}, \texttt{lr})\end{tabular} & \rev{(200, Adam, 1e-3)} & \rev{(200, Adam, 1e-4)} & \rev{(100, Adam, 1e-4)} & \rev{(200, SGD, 1e-2)} & \rev{(100, Adam, 1e-4)} & \rev{(200, Adam, 1e-4)} & \rev{(200, Adam, 1e-4)} \\ \hdashline
\rev{Overall} & \rev{4.65\% ± 0.00\%} & \rev{24.05\% ± 0.34\%} & \rev{16.62\% ± 0.86\%} & \rev{5.16\% ± 0.20\%} & \rev{14.42\% ± 0.13\%} & \rev{17.27\% ± 0.38\%} & \rev{12.07\% ± 0.28\%} \\
\rev{Type 1} & \rev{3.51\% ± 1.24\%} & \rev{21.71\% ± 0.47\%} & \rev{22.55\% ± 0.80\%} & \rev{4.80\% ± 0.38\%} & \rev{--} & \rev{15.05\% ± 0.35\%} & \rev{7.85\% ± 0.36\%} \\
\rev{Type 2} & \rev{6.92\% ± 0.89\%} & \rev{21.06\% ± 0.50\%} & \rev{18.57\% ± 0.51\%} & \rev{4.11\% ± 0.38\%} & \rev{--} & \rev{16.43\% ± 0.40\%} & \rev{9.81\% ± 0.27\%} \\
\rev{Type 3} & \rev{0.90\% ± 1.27\%} & \rev{23.72\% ± 0.68\%} & \rev{14.39\% ± 1.34\%} & \rev{5.21\% ± 0.42\%} & \rev{17.18\% ± 0.15\%} & \rev{--} & \rev{13.23\% ± 0.54\%} \\
\rev{Type 4} & \rev{8.33\% ± 1.18\%} & \rev{28.56\% ± 0.93\%} & \rev{15.32\% ± 0.61\%} & \rev{6.62\% ± 0.52\%} & \rev{13.16\% ± 0.45\%} & \rev{--} & \rev{20.22\% ± 0.30\%} \\
\rev{Type 5} & \rev{1.45\% ± 2.05\%} & \rev{32.09\% ± 0.77\%} & \rev{19.43\% ± 1.53\%} & \rev{8.12\% ± 0.90\%} & \rev{12.23\% ± 0.47\%} & \rev{25.70\% ± 0.50\%} & \rev{--} \\
\rev{Type 6} & \rev{4.55\% ± 0.00\%} & \rev{25.97\% ± 1.06\%} & \rev{15.32\% ± 2.37\%} & \rev{8.00\% ± 1.08\%} & \rev{9.29\% ± 0.22\%} & \rev{18.11\% ± 0.45\%} & \rev{--} \\
\midrule\midrule
\rev{Groh et al.~\cite{groh2021evaluating}: Overall} & \rev{26.7\%} & \rev{20.2\%} & \rev{27.4\%} & \rev{11.4\%} & \rev{13.8\%} & \rev{13.4\%} & \rev{7.7\%} \\

\bottomrule
\end{tabular}%
}
\caption{
Benchmark results (3 repeated runs; mean $\pm$ std. dev.) of \fitzdsetC for all the experiments originally reported by Groh et al.~\cite{groh2021evaluating}.
The metrics being reported are the overall accuracy (``Overall'') and FST-specific accuracy (``Type $x$'' corresponds to Fitzpatrick skin tone $x$). 
\textbf{Verified}: the models are tested on a set of 215 images that were verified to be diagnostic of the disease label by a board-certified dermatologist. 
\textbf{Random (Stratified)}: the test partition contains 20\% of the dataset, randomly sampled stratified on the disease labels. \textbf{Source \{A, B\}}: the models were tested on all the images from Atlas Dermatologico and DermaAmin respectively. \textbf{FST {$\bm{xx-yy}$}}: the models were tested on images with FST labels $xx, \cdots, yy$. For all the experiments, the training and the validation partitions were drawn from the remaining images from \fitzdsetC. \rev{It is important to note that the results in the last row have been reported verbatim from Groh et al.~\cite{groh2021evaluating}, whose training and evaluation partitions differ significantly from our work, and therefore these metrics are not directly comparable.}
}
\label{tab:f17k_results}
\end{table}

\subsubsection{Erroneous images}
In a recent study, Pakzad et al.~\cite{pakzad2022circle} reported the presence of erroneous or outlier non-skin images in \fitzdset. Using an outlier detection approach based on distance to the nearest neighbors in the embedding space, we rank images in the dataset based on their probability of being an outlier.
Sample outliers, as shown in Fig.~\ref{fig:fitz_outliers}, include non-dermatological imaging modalities (\eg histopathology, radiology, microscopy, fundus), images of plants (leaves, trees) and animals (\eg rodents, bugs, poultry), etc. 
Worryingly, \fitzdset does not contain information regarding which images are non-dermatological, which consequently impacts the training and evaluation of models.

\begin{table}[!ht]
\centering
\resizebox{\textwidth}{!}{%
\setlength{\tabcolsep}{0.7em}
\def\arraystretch{1.25}
\begin{tabular}{lccccccc}
\toprule
\begin{tabular}[l]{@{}l@{}}\textbf{Hyperparameters}\\\textbf{optimized on $\rightarrow$}\end{tabular} & Verified & \begin{tabular}[c]{@{}c@{}}Random\\(Stratified)\end{tabular} & \begin{tabular}[c]{@{}c@{}}Source A\\(Atl. Derm.)\end{tabular} & \begin{tabular}[c]{@{}c@{}}Source B\\(DermaAmin)\end{tabular} & FST 3--6 & FST 1--2 \& 5--6 & FST 1--4 \\
\hdashline
\begin{tabular}[l]{@{}l@{}}Best\\Hyperparameters\\(\texttt{n\_epochs}, \texttt{optim}, \texttt{lr})\end{tabular} & \rev{(200, Adam, 1e-3)} & \rev{(200, Adam, 1e-4)} & \rev{(100, Adam, 1e-4)} & \rev{(200, SGD, 1e-2)} & \rev{(100, Adam, 1e-4)} & \rev{(200, Adam, 1e-4)} & \rev{(200, Adam, 1e-4)} \\ \hdashline
\begin{tabular}[l]{@{}l@{}}\textbf{Holdout set performance}\\\textbf{(overall test accuracy)}\end{tabular} & & & & & & & \\
\midrule
\rev{Verified} & \rev{\hphantom{0}4.65\% ± 0.00\%} & \rev{\hphantom{0}4.34\% ± 0.22\%} & \rev{\hphantom{0}4.50\% ± 0.22\%} & \rev{\hphantom{0}3.88\% ± 0.96\%} & \rev{\hphantom{0}4.50\% ± 0.22\%} & \rev{\hphantom{0}4.34\% ± 0.22\%} & \rev{\hphantom{0}4.34\% ± 0.22\%} \\
\rev{Random Holdout} & \rev{20.20\% ± 0.32\%} & \rev{24.05\% ± 0.34\%} & \rev{23.23\% ± 0.43\%} & \rev{19.25\% ± 0.57\%} & \rev{23.23\% ± 0.43\%} & \rev{24.05\% ± 0.34\%} & \rev{24.05\% ± 0.34\%} \\
\rev{Source A} & \rev{14.96\% ± 0.77\%} & \rev{16.38\% ± 0.77\%} & \rev{16.62\% ± 0.86\%} & \rev{14.28\% ± 0.30\%} & \rev{16.62\% ± 0.86\%} & \rev{16.38\% ± 0.77\%} & \rev{16.38\% ± 0.77\%} \\
\rev{Source B} & \rev{\hphantom{0}4.78\% ± 0.20\%} & \rev{\hphantom{0}4.50\% ± 0.32\%} & \rev{\hphantom{0}4.50\% ± 0.32\%} & \rev{\hphantom{0}5.16\% ± 0.20\%} & \rev{\hphantom{0}4.50\% ± 0.32\%} & \rev{\hphantom{0}4.50\% ± 0.32\%} & \rev{\hphantom{0}4.50\% ± 0.32\%} \\
\rev{FST 3--6} & \rev{12.65\% ± 0.23\%} & \rev{14.36\% ± 0.15\%} & \rev{14.42\% ± 0.13\%} & \rev{11.80\% ± 0.69\%} & \rev{14.42\% ± 0.13\%} & \rev{14.36\% ± 0.15\%} & \rev{14.36\% ± 0.15\%} \\
\rev{FST 1--2 \& 5--6} & \rev{14.56\% ± 0.53\%} & \rev{17.27\% ± 0.38\%} & \rev{16.99\% ± 0.33\%} & \rev{13.78\% ± 0.41\%} & \rev{16.99\% ± 0.33\%} & \rev{17.27\% ± 0.38\%} & \rev{17.27\% ± 0.38\%} \\
\rev{FST 1--4} & \rev{11.34\% ± 0.66\%} & \rev{12.07\% ± 0.28\%} & \rev{11.92\% ± 0.38\%} & \rev{\hphantom{0}8.08\% ± 4.58\%} & \rev{11.92\% ± 0.38\%} & \rev{12.07\% ± 0.28\%} & \rev{12.07\% ± 0.28\%} \\

\bottomrule
\end{tabular}%
}
\caption{\rev{
Understanding how \fitzdsetC classification performance varies with change in hyperparameters. The columns denote the optimal hyperparameters for each of the seven experimental settings, and the rows represent the overall test accuracies for all the seven settings when models trained with those particular hyperparameters are evaluated. For example, when a model trained with \saynew{Source A}'s optimal hyperparameters (i.e., 100 training epochs, Adam optimizer, $1e-4$ learning rate) is used to generate predictions for the \saynew{Verified} setting, the overall test accuracy is $4.50\% \pm 0.22\%$.
}
}
\label{tab:f17k_hparam_7x7}
\end{table}

\subsubsection{Non-standardized data partitioning}

The \fitzdset benchmarks by Groh et al.~\cite{groh2021evaluating}, \revnewer{as well as several works that followed~\cite{du2022fairdisco,aayushman2024fair,ghadiri2024xtranprune,wang2024achieving}}, also suffer from another major problem: the lack of a strictly held-out test partition.
For all their skin condition prediction experiments, the authors only used a training and a validation set, and used the terms \say{validation} and \say{testing} interchangeably in the paper. This can also be verified in their accompanying code implementation, where the data partitions used to select the best epoch during training~\cite{fitzpatrick17k_github_l82}
(\say{\textit{the epoch with the lowest loss on the validation set}}) and to report the final results~\cite{fitzpatrick17k_github_l342}
are the same. This violates the 
\ka{fundamental rules}
of machine learning model training and evaluation, where the validation and the testing partitions must be separate disjoint sets, and the former is used for choosing the best performing model during training and hyperparameter selection, while the latter is reserved only for the final model evaluation and is never used during training.

\subsubsection{\rev{Correcting} \fitzdset}

\begin{figure}[!ht]
     \centering
     \begin{subfigure}[b]{0.47\textwidth}
         \centering
         \includegraphics[width=\textwidth]{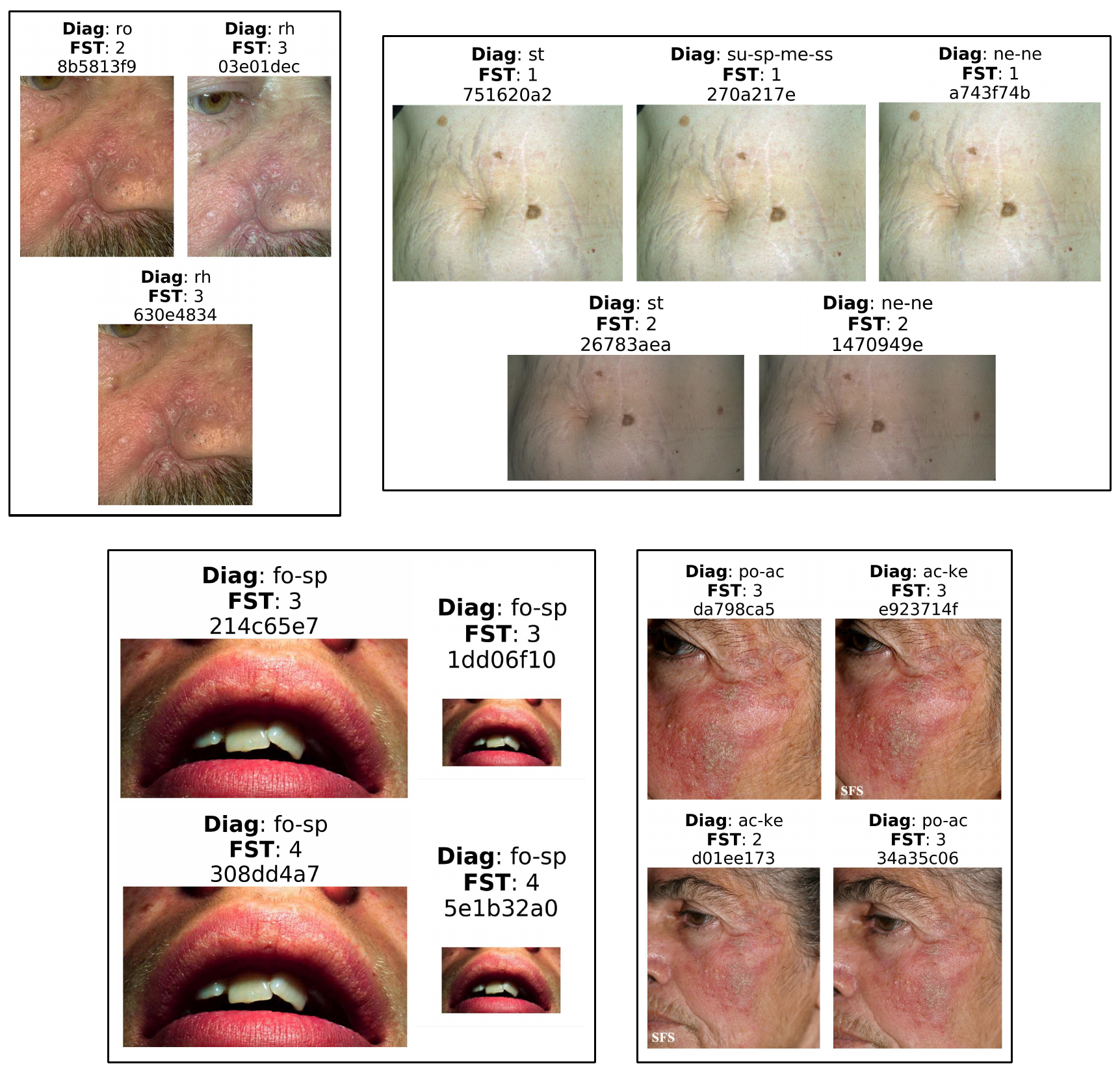}
         \caption{\rev{Sample \saynew{heterogenous} clusters of duplicate images.}}
     \end{subfigure}
     \hfill   
     \begin{subfigure}[b]{0.47\textwidth}
         \centering
         \includegraphics[width=\textwidth]{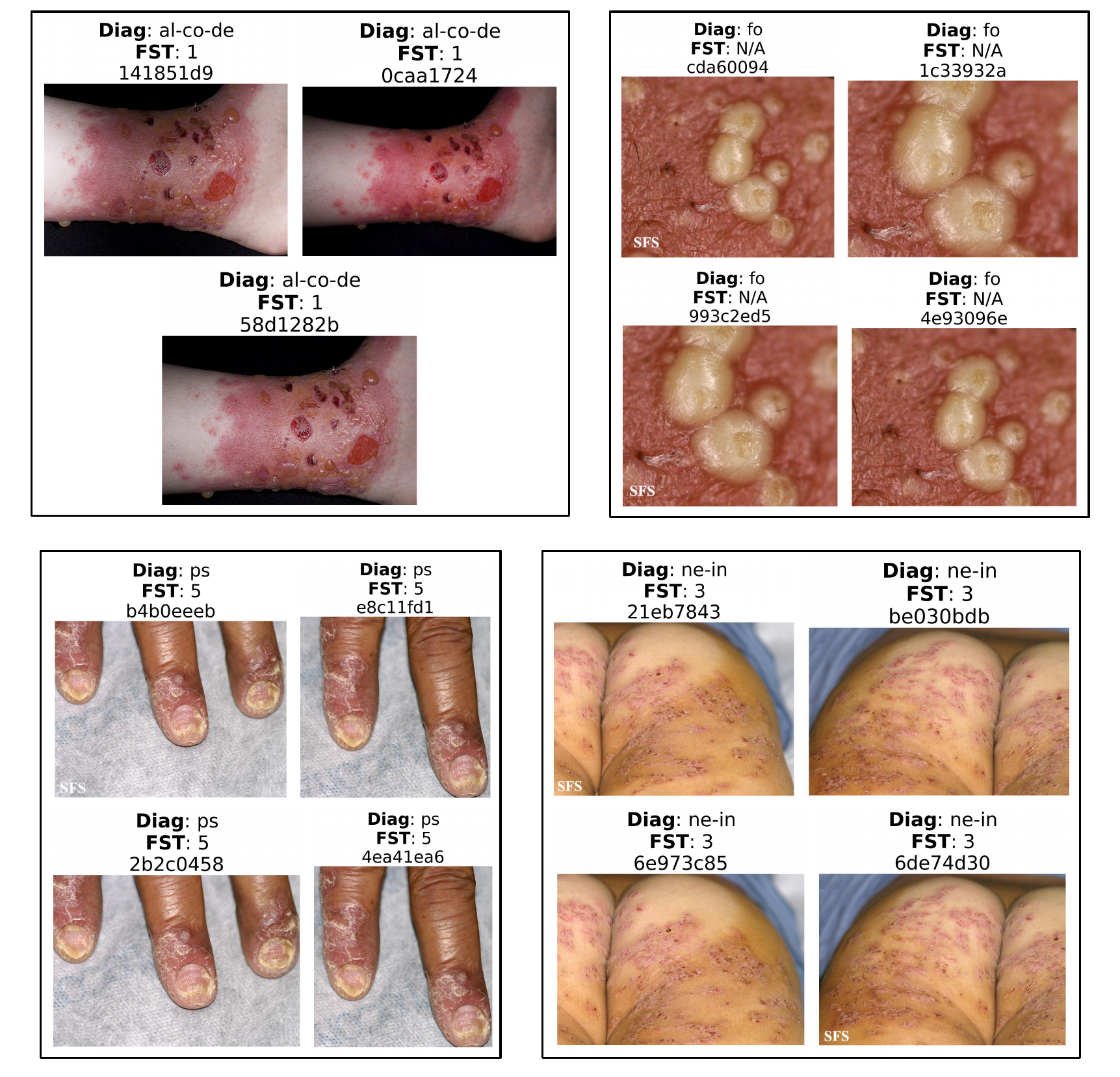}
         \caption{
         \rev{
         Sample \saynew{homogenous} clusters of duplicate images.
         }
         }
     \end{subfigure}

     \caption{
         \rev{
         Visualizing sample (a) \saynew{heterogenous} and (b) \saynew{homogenous} clusters of duplicate images. A cluster is considered \saynew{homogenous} if all the duplicate images in it share the same diagnosis and FST labels; otherwise it is deemed \saynew{heterogenous}.
         }
     }
     \label{fig:homogenous_vs_hetergenous}
\end{figure}

In light of the numerous aforementioned issues with \fitzdset: data duplication, conflicting labels, the presence of erroneous images, and the absence of a well-defined test partition, we attempt to clean up \fitzdset and present a smaller, yet more reliable dataset. Specifically, we remove clusters of duplicates (this includes duplicate pairs), keeping one image from each cluster if there are no conflicting diagnosis or FST labels within the cluster \rev{, i.e., a \say{homogenous cluster} (Fig.~\ref{fig:homogenous_vs_hetergenous})}. Next, we remove the erroneous images from the dataset and refer to this \say{cleaned} version of \fitzdset as \fitzdsetC.

In the absence of standardized dataset partitions, researchers who used \fitzdset for their models had to resort to generating their own splits~\cite{abid2022meaningfully,du2022fairdisco,pakzad2022circle}, making it very hard for models across papers to be compared. To resolve this, we present standardized training, validation, and testing partitions for \fitzdsetC for the skin image analysis community to use, obtained by splitting \fitzdsetC in the ratio of 70:10:20, stratified on the diagnosis labels. \ka{Table~\ref{tab:datasets_summary} summarizes the two datasets: \fitzdset and \fitzdsetC, listing the number of images in their respective partitions.}

\begin{table}[!ht]
\centering
\resizebox{0.9\textwidth}{!}{%
\setlength{\tabcolsep}{0.7em}
\def\arraystretch{1.25}
\begin{tabular}{@{}cc:cc:ccc@{}}
\toprule
\textbf{Dataset} & \textbf{\begin{tabular}[c]{@{}c@{}}Brief\\ Description\end{tabular}}                                                                                                             & \textbf{\#Images} & \textbf{\#Diagnoses} & \textbf{\begin{tabular}[c]{@{}c@{}}\#Training\\ Images\end{tabular}} & \textbf{\begin{tabular}[c]{@{}c@{}}\#Validation\\ Images\end{tabular}} & \textbf{\begin{tabular}[c]{@{}c@{}}\#Testing\\ Images\end{tabular}} \\ \midrule
\dermamnist      & \begin{tabular}[c]{@{}c@{}}The original\\ \dermamnist dataset.\end{tabular}                                                                                                      & 10,015            & 7                   & 7,007                                                                & 1,003                                                                  & 2,005                                                               \\ \hdashline 
\dermaC          & \begin{tabular}[c]{@{}c@{}}The \say{corrected} version of\\ \dermamnist, without any\\ data leakage.\end{tabular}                                                                    & 10,015            & 7                   & \rev{8,215}                                                                & \rev{573}                                                                   & \rev{1,227}                                                               \\ \hdashline 
\dermaE          & \begin{tabular}[c]{@{}c@{}}The \say{extended} version of \\ \dermamnist, without any \\ data leakage and with \\ more images.\end{tabular}                                           & \rev{11,719}            & 7                   & 10,015                                                               & 193                                                                   & \rev{1,511}                                                               \\ \midrule
\fitzdset        & \begin{tabular}[c]{@{}c@{}}The original \\ \fitzdset dataset.\end{tabular}                                                                                                       & 16,577            & 114                 & 12,751                                                               & 3,826                                                                  & 0                                                                  \\ \hdashline 
\fitzdsetC       & \begin{tabular}[c]{@{}c@{}}The \say{cleaned} version of \\ \fitzdset, with standardized\\ train-valid-test splits after\\  removing duplicates and\\  erroneous images.\end{tabular} & 11,394            & 114                 & 7,975                                                                & 1,139                                                                  & 2,280                                                               \\ \bottomrule
\end{tabular}%
}
\caption{
\ka{
Summary statistics for the two datasets analyzed in this paper and their corresponding corrected versions proposed. For both \fitzdset and \fitzdsetC, the partitions correspond to the experiment titled \saynew{Random}, both in Table~\ref{tab:f17k_results} and in Groh et al.~\cite{groh2021evaluating}.
}
}
\label{tab:datasets_summary}
\end{table}

Finally, we also provide benchmarks for \fitzdsetC using all the different experimental settings proposed by Groh et al.~\cite{groh2021evaluating} in Table~\ref{tab:f17k_results}.
\rev{We perform a hyperparameter search for each experimental setting over the space of optimizers (\{Adam, SGD\}), learning rate (\{$1e-2$, $1e-3$, $1e-4$\}), and number of training epochs (\{20, 50, 100, 200\}), and list the number of images in the training, validation, and testing partitions. For added robustness, we repeat each experiment using 3 random seeds.
We also observed that using one setting's optimal hyperparameter choices to evaluate another setting's test partition does not considerably degrade the classification performance (Table~\ref{tab:f17k_hparam_7x7}).
}

\section{Discussion}

In this paper, we examine the data quality of 
\rev{three}
popular and large skin image analysis datasets: \dermamnist from the \medmnist dataset 
\rev{and its source \ham dataset}
(\hamcount{} dermoscopic images of skin lesions) and \fitzdset (\fitzcount{} clinical images of skin diseases). For \dermamnist, we investigate the extent of data leakage across its training, validation, and testing partitions, and propose corrected (\dermaC) and extended (\dermaE) versions. We conducted benchmark evaluations using multiple methods and compare the results to those of \dermamnist across all datasets. For \fitzdset, we perform a systematic analysis encompassing data duplication, mislabeling of diagnosis and Fitzpatrick skin tone labels, identification of erroneous images, as well as highlighting the use of non-standard data partitions. Finally, we propose a cleaned version of the dataset with standardized partitions called \fitzdsetC, and release the corresponding updated benchmarks.

\subsection{\dermamnist}

\subsubsection{Data leakage and benchmarks}

The extent of data leakage in \dermamnist emanating from improper data partitioning is quite severe, with 1,006 of the 7,470 unique lesions ($\sim$13.47\%) in the dataset being present in more than 1 partition (Fig.~\ref{fig:dermamnist} (c)). Patient-level stratification, or in this case, lesion-level stratification, is crucial to ensure that the model does not \say{see} the lesions in the held-out test set while training. Consequently, the lack of such a stratification implies that the corresponding benchmark classification results are not truly reflective of the models' generalization capability and, therefore, we strongly suspect that the models' performance could be inflated. \rev{On the other hand, since most works use the entire \ham dataset for training instead of splitting it like in the case of \dermamnist, undetected duplicates in \ham are arguably a less severe issue.}

However, a word of caution on comparing the benchmark results of \dermamnist and \dermaC (Table~\ref{tab:dermamnist}): the two datasets do not share the same test partitions. Specifically, \dermaC's test set was obtained from \dermamnist's test set by removing all images whose \texttt{lesion ID}s were present in the test set, and therefore, the former is a subset of the other (i.e., \dermaC test set $\subset$ \dermamnist test set). Similarly, as shown in Fig.~\ref{fig:dermamnist} (c), the training set of \dermaC is larger than that of \dermamnist. For these reasons, contrary to the assumption that fixing data leakage in \dermamnist should decrease performance benchmarks, we emphasize that performance benchmarks for \dermamnist and \dermaC should not be compared since these models have been trained and evaluated on dissimilar partitions.

\subsubsection{A more challenging dataset}
\ka{While \ham is indeed a valuable dataset for skin image analysis research, models trained on \ham do not necessarily perform well when evaluated on other skin lesion image datasets~\cite{yoon2019generalizable}.}
This was also observed by the organizers of the ISIC 2018 Challenge~\cite{codella2019skin} where they relied on external testing data for ranking the submissions (\say{\textit{multipartition test sets containing data not reflected in training dataset are an effective way to differentiate the ability of algorithms to generalize}}). Therefore, following the ISIC 2018 Challenge, we use the images from the Challenge's validation and testing partitions, resized to \lowres and \hires, to create the validation and testing partitions, respectively, of the newly proposed \dermaE dataset, thus allowing for a more robust assessment of the skin lesion diagnosis models trained on this dataset.

\subsubsection{Incorrect scaling of images}
Another issue with the \dermamnist benchmarks was the use of \lowres images to report the results on \hires resolutions. As we show in 
Fig.~\ref{fig:dermamnist_part2},
the information lost when downsampling from $600 \times 450$ to \lowres is quite significant, and it is impossible to recover this when upsampling from \lowres to \hires. This can also be observed in the quantitative results (Table~\ref{tab:dermamnist}): intuitively, given that we ensure that the models do not overfit, we would expect that a model with larger capacity (ResNet-50) trained on high\rev{-} resolution images (\hires) should perform better than a lower capacity model (ResNet-18) trained on lower resolution images (\lowres). However, this is not the case with \dermamnist results, where ResNet-18/\lowres models perform [AUC; ACC] better than ResNet-50/\hires models: [0.917; 0.735] versus [0.912; 0.731]. On the other hand, with \dermaC and \dermaE, ResNet-50/\hires models do perform better than their ResNet-18/\lowres counterparts.

\subsection{\fitzdset}

\subsubsection{Duplicate detection and guarantees}
Because of the large scale of the \fitzdset dataset (\fitzcount{} images), manual review of images to verify duplicates is virtually impossible due to the huge combinatorial space: there are $\binom{\fitzcount{}}{2} = 137$ million pairs of images. Even worse, if we want to verify triplicates (i.e., 3 images that are copies of one another), we would have to review $\binom{\fitzcount{}}{3} = 759$ billion clusters of 3 images, and the number keeps growing as the size of the clusters being reviewed increases.
Therefore, we rely on automated methods for duplicate detection followed by a manual review of duplicates above a reasonable similarity threshold, and our manual review results confirmed a near-perfect agreement with the algorithm's results. Our experiments showed that using a second method of duplicate detection (\pypackage{cleanvision}) was able to discover an additional 19 pairs of duplicates. These pairs were also detected by \pypackage{fastdup}, but their similarity scores fell just short of the chosen thresholds of 0.90 and 0.95, and therefore were absent from our manual review of \pypackage{fastdup}'s results. Our dataset cleaning pipeline,
\ka{publicly available on GitHub~\cite{critique_github_cite}},
is highly modular and configurable, allowing users to adjust similarity thresholds to exclude and use multiple duplicate lists (\pypackage{fastdup} and \pypackage{cleanvision}) for cleaning, decide whether they want to remove duplicate clusters altogether or retain one representative image from each cluster, decide whether they want to remove images with unknown FST labels, and decide which outliers to exclude based on a similarity threshold. While we can claim with a high degree of certainty that the new cleaned dataset \fitzdsetC will be devoid of any duplicates, the large scale of manual review required makes it nearly impossible to guarantee it.

\subsubsection{Dealing with conflicting diagnosis and FST labels}
Several duplicates in \fitzdset, despite being near identical copies, do not share the same diagnosis labels (Fig.~\ref{fig:fitz_duplicates} (a, b)). This affects the training and evaluation of models, since a model can be incorrectly penalized for its prediction because of the conflicting labels. However, correcting these labels, so that copies of the same image have the same diagnosis label, requires a domain expert (i.e., a dermatologist) to go through all the images and confirm and correct their labels. Unfortunately, even if such an endeavor were to be undertaken, these diagnoses will not be histopathology-confirmed and the accuracy of diagnoses confirmed through images alone (a scenario similar to \say{store-and-forward} teledermatology) is expected to be lower than those confirmed through in-person patient visits~\cite{s2009teledermatology,gerhardt2021assessing,gunn2020asynchronous}.

Another approach to resolving the diagnosis label conflicts could be through mapping the diagnoses in \fitzdset to the \rev{World Health Organization (WHO)} International Classification of Diseases, Eleventh Revision (ICD-11)~\cite{who2019icd11}, and if multiple diagnoses belong to the same \say{parent}, their label conflict can be resolved by assigning the \say{parent}'s label to both. However, when mapping the diagnoses in \fitzdset to using the ICD-11 Browser~\cite{whoICD11Mortality}
we ran into the following issues:

\begin{itemize}
    \item Several diagnosis labels did not yield any matches (e.g., \say{acquired autoimmune bullous diseaseherpes gestationis}, \say{nematode infection}, \say{neurotic excoriations}, \say{pediculosis lids}).
    \item There were diagnosis labels where one of the labels has an entry in ICD-11, but other seemingly related labels do not (e.g., \say{basal cell carcinoma} exists in ICD-11, but possibly related \say{basal cell carcinoma morpheiform} and \say{solid cystic basal cell carcinoma} do not).
    \item There were labels where we found near but not exact matches (e.g., the label \say{erythema annulare centrifigum} in \fitzdset does not have an exact match, but ICD-11 contains \say{Erythema annulare}; other partial matching [\fitzdset; ICD-11] labels include [\say{hidradenitis}; \say{Hidradenitis suppurativa}], [\say{fixed eruptions}, \say{Fixed drug eruption}], [\say{lupus subacute}, \say{Subacute cutaneous lupus erythematosus}], [\say{porokeratosis actinic}, \say{Disseminated superficial porokeratosis actinic}], etc.
\end{itemize}

\rev{Overall, 8 diagnosis labels had no matches in ICD-11. An additional 29 diagnoses yielded partial matches to entries in ICD-11, and therefore could not be reliably mapped to a single entry. For these diagnoses, i.e., those that had partial or no matches in ICD-11, we expanded our search to the ICD-11 Coding Tool Mortality and Morbidity Statistics (MMS)~\cite{whoICD11Coding}
and the ICD-11 Classification of Dermatological Diseases~\cite{white2022who}~\cite{whoICD11Derm},
but this did not resolve any issues.}

\rev{Additionally, we also carried out a similar diagnosis lookup on the Systematized Medical Nomenclature of Medicine - Clinical Terminology (SNOMED-CT), a comprehensive clinical global healthcare terminology that may better represent medical vocabulary than ICD-11~\cite{thun2018icd}. Although we were able to find more matches to SNOMED-CT Identifiers~\cite{ihtsdotoolsSNOMEDInternational} (SCTIDs) than to entries in ICD-11 and its derivatives, it was still not perfect: we ended up with 17 and 3 diagnosis labels with partial and no matches, respectively. We make the results of our \fitzdset diagnosis-to-\{ICD-11 classification code, SCTID\} mapping publicly available online.}

As a recommendation for the future datasets containing clinical images of skin diseases, we believe that adding \rev{either} ICD classification codes \rev{or SNOMED-CT Identifiers (SCTIDs)} for the disease labels would be a helpful addition to the metadata, and would greatly enhance the usability of such datasets for hierarchical diagnosis methods~\cite{barata2019deep,benyahia2022hierarchical,yu2022skin}.

Similar conflicts were observed for FST labels of images, where images with 
\rev{very}
high similarity varied in their FST labels, sometimes by as much as 4 tones (Fig.~\ref{fig:fitz_duplicates} (a, b), Fig.~\ref{fig:fitzpatrick_stats}). A potential solution to this would be to re-assess the images with conflicting FST labels. This would involve either obtaining manual and verified healthy skin segmentation masks or using an automated healthy skin segmentation method, followed by mapping the skin tone estimated using these healthy skin pixels to an FST label. However, as noted by Groh et al.~\cite{groh2021evaluating}, collecting manual segmentation masks is expensive, and automated skin segmentation approaches suffer from their own set of challenges~\cite{naji2019survey, nanni2023standardized}, including but not limited to susceptibility to non-standardized illumination (Fig.~\ref{fig:fitz_duplicates} (d), Fig.~\ref{fig:fitz_duplicates_part2} (c)), the presence of multiple objects (Figs.~\ref{fig:fitz_duplicates_part2} (b, c)), and/or low quality images (Fig.~\ref{fig:fitz_duplicates} (e), Fig.~\ref{fig:fitz_duplicates_part2} (d)). The presence of diffuse pathologies such as \say{acne}, \say{acne vulgaris}, and \say{disseminated actinic porokeratosis}, where the diseased regions of the skin are spread over a wide region and the exact boundaries between diseased and healthy regions are not well-defined, further compounds the task for any automated healthy skin detection approach.

Because of these intractable issues, we make the decision of removing duplicate images that have conflicting diagnosis or FST labels. While we realize that relabeling might be able to solve these label conflicts, we also acknowledge the enormous time, effort, and expenses such an endeavor would entail.

\subsubsection{Updated benchmarks}
As discussed above, Groh et al.~\cite{groh2021evaluating} used non-standardized dataset partitions for training and evaluating their models for benchmarking, the most notable issue being the absence of a well-defined and held-out test set, thus violating 
\ka{one of the most basic rules of machine learning.}
The use of the same dataset partition for validation, i.e., choosing the best performing models, and for testing ensures that the reported performance is the best possible, but it is a poor reflection of the models' generalization capabilities, since the models have been overfit to the testing data. This issue, coupled with the presence of duplicates in \fitzdset, which would inevitably lead to data leakage across partitions, is the reason for the difference between \fitzdset model benchmarks by Groh et al. and our \fitzdsetC 
benchmarks (Table~\ref{tab:f17k_results}). For the \say{Verified} experiments (i.e., \say{\textit{testing on the subset of images labeled by a board-certified dermatologist as diagnostic of the labeled condition and training on the rest of the data}}~\cite{groh2021evaluating}), the number of training images for \fitzdset decreased from 16,229 to 10,060, meaning the models were trained on considerably fewer images. On the other hand, the number of testing images in \fitzdsetC decreased by $\sim$38\% (from 348 in \fitzdset to 215) after dataset cleaning, implying that 133 images were duplicates in \fitzdset's testing partition, which explains why their benchmark results were inflated. Similar trends can be observed across all the benchmark experiments: the number of training images for \say{Source A}, \say{Source B}, and \say{Fitz 1-2 \& 5-6} experiments nearly halved after dataset cleaning, and the number of both training and testing images for the \say{Fitz 1-4} experiments decreased by more than 50\%.

We believe that the newly proposed \fitzdsetC along with the well-defined and disjoint training, validation, and testing partitions, the benchmarks, and the publicly available training and evaluation code will help other researchers better utilize the dataset and make comparison across methods easier and standardized.

\subsection{Comparison to other works}

One of the first works on analyzing skin image analysis datasets by Abhishek~\cite{abhishek2020input} focused on ISIC Challenge datasets for the skin lesion segmentation task, specifically on the challenges from 2016, 2017, and 2018. They found these datasets to have considerable overlap across their training partitions, and this can be attributed to all of them being subsets of the ISIC Archive~\cite{ISICArchive}. Their analysis, however, was limited to overlap detection simply based on the filenames, since all the filenames in the ISIC Archive follow the template \texttt{ISIC\_$\langle$ISIC identifier$\rangle$.jpg}.
Specific to \fitzdset, Pakzad et al.'s work~\cite{pakzad2022circle} was the first to mention the presence of \say{erroneous and wrongly labeled images}, and how the images therein have non-standard illumination and camera perspectives, highlighting the need for cleaning \fitzdset.
Groger et al.~\cite{groger2023reliable} conducted a data quality analysis on the presence of \say{irrelevant samples}, near duplicates, and label errors in 6 datasets. They used a self-supervised DL-based method for generating rankings of images potentially containing data quality issues, following which, 3 experts including a board-certified dermatologist manually reviewed the images and answered a questionnaire. The authors also noted that non-expert annotations may be sufficient for confirming duplicate images.
Vega et al.~\cite{vega2023analysis} analyzed a popular and well-cited monkeypox skin image dataset, and found it to contain \say{medically irrelevant images}. They discovered that the images were extracted from online repositories through web-scraping and lacked medical validation. Finally, their experiments showed that the claims made by the dataset's authors about the utility of the images might not have been true, since a model trained by Vega et al. on \say{blinded} images (i.e., images where the regions of interest related to the disease were covered by black rectangles) was also able to accurately classify the diseases.

The closest dataset analysis to our work, both in the scale of the analyses and the sizes of the datasets analyzed, is the work by Cassidy et al.~\cite{cassidy2022analysis} on the ISIC Challenge datasets from 2016 through 2020, where the authors employed a multi-step duplicate removal strategy. First, similar to 
our previous work~\cite{abhishek2020input}, they removed duplicates across dataset partitions based on filenames. Next, they used the following tools and methods on \say{\textit{a random selection of training images for 72 hours}}: (a) a Python library called \texttt{ImageHash} to measure similarity based on several hashing methods (average, perceptual, difference, and wavelet), (b) mean-squared error (MSE)-based image similarity detection, (c) structural similarity index measure (SSIM)-based image similarity detection, and (d) cosine similarity-based detection. Finally, they used a Python library called \texttt{FSlint} to detect duplicates based on MD5 and SHA-1 file checksum signatures.

The analyses of the 
\rev{three}
datasets in this paper: \dermamnist,
\rev{\ham,}
and \fitzdset arguably goes beyond those presented in these previous works. For \dermamnist, since we were able to request the exact training-validation-testing splits' filenames, we could simply cross-reference those to the publicly available \ham metadata to detect and correct data leakage. 
Our analysis of \fitzdset is much more 
\rev{in-depth}
and involved: we start with finding similar images and erroneous images in the embeddings' latent space, followed by a manual non-expert review to confirm the duplicates, and then coalesce duplicates to form larger clusters wherever applicable \rev{(Fig.~\ref{fig:f17k_coalesce_duplicates})}.  
The filenames in \fitzdset follow the template \texttt{$\langle$MD5\_hash$\rangle$.jpg}, meaning that all the images have a unique MD5 checksum, which in turn implies that file checksum-based duplicate detection would not be possible.
The Python library \pypackage{cleanvision} that we use for duplicate detection already relies upon \texttt{ImageHash} for hashing-based similarity measurement.
Measuring similarity scores in the embedding space is arguably superior to doing so in the input space (e.g., MSE, SSIM-based duplication checks by Cassidy et al.~\cite{cassidy2022analysis}), since a latent representation goes beyond pixel-level information and is able to better capture semantic information about the image as a whole.
Finally, because of the large scale of \fitzdset (\fitzcount{} images), manual review by experts become cost prohibitive.

\subsection{Data leakage and reproducibility crisis}

Kapoor et al.~\cite{kapoor2023leakage} presented 
\rev{an in-depth}
review of instances of data leakage in machine learning (ML) applications across 17 scientific domains, and how such leakage can lead to a reproducibility crisis in ML-based science. 
They also proposed a hierarchical taxonomy of leakage types, which has 8 types of leakage across 2 levels. Our analysis of \dermamnist shows that it suffers from \say{\textbf{[L1.4]}: \emph{Duplicates in datasets}} and \say{\textbf{[L3.2]}: \emph{Non-independence between train and test samples}}, where the latter is a direct consequence of the former. Concerningly, \fitzdset exhibited three types of data leakage: \say{\textbf{[L1.1]}: \emph{No test set}}, since it did not have a disjoint held-out test set and had duplicate images, along with the aforementioned \textbf{[L1.4]} and \textbf{[L3.2]}.

\medskip
\par
In conclusion, in this paper, we examined the data quality of 
\rev{three}
popular skin image analysis datasets: \dermamnist from the \medmnist collection,
\rev{its source \ham}, 
and \fitzdset. For \dermamnist, we investigated the extent of data leakage across its partitions and proposed 
\rev{two new and improved versions: a corrected dataset that does not have any leakage (\dermaC) and an extended and arguably more challenging dataset (\dermaE) that is almost the same as the ISIC 2018 Challenge dataset except the images are resized to \lowres and \hires and the \say{easter egg} image is removed from the test partition.}
\rev{We also investigated the presence of duplicates in \ham, the source dataset for \dermamnist, and discovered 18 new duplicate image pairs that were unaccounted for in the metadata.}
For \fitzdset, we conducted a systematic analysis encompassing data duplication, mislabeling of diagnosis and skin tone labels, as well as the identification of outlier images, followed by cleaning the dataset to propose a cleaned version \fitzdsetC \rev{with standardized training, validation, and testing partitions}. 
\rev{We also showed how \fitzdset contains labels that do not fully align, which we recommend they should, with internationally recognized standards such as ICD-11 and SNOMED-CT, limiting its utility for hierarchical and differential diagnosis-based approaches.} For all the datasets, we conducted benchmark evaluations using multiple methods, repeated 3 times for robustness.
The primary objective of this paper is to raise awareness about potential data quality issues that may arise in large datasets, and how these issues can go unnoticed even in popular datasets, casting doubts on the conclusions made about the robustness and the generalizability of the models trained on these datasets. We hope this can serve as a call to action for more stringent data quality assessments. To facilitate this, we plan to make our evaluation method and code publicly available so that they can be used and extended upon for evaluating more existing and new datasets.

\section{Methods}

\subsection{\dermamnist}

\subsubsection{Detecting and correcting data leakage}
We used \ham's publicly available metadata, which contains \texttt{image ID} to \texttt{lesion ID} mappings, and \dermamnist partitions' filenames, which contain \texttt{image ID}s, to perform an inner-join using the Python library \texttt{pandas}. By doing so, we were able to obtain the list of images belonging to the same lesion that were common across train-valid, train-test, valid-test, and train-valid-test partitions.

To correct the leakage, we follow this approach: if images of a \texttt{lesion ID} exist in the training partition, move all images belonging to that \texttt{lesion ID} from validation and testing partitions to the training. For instance, for the images visualized in Fig.~\ref{fig:dermamnist} (a), since images belonging to lesion ID \texttt{HAM\_0002364}, i.e., \texttt{ISIC\_0024712}, \texttt{ISIC\_0025446}, and \texttt{ISIC\_0030348}, are present in the training partition, we move the other images of the same lesion, i.e., \texttt{ISIC\_0029838} and \texttt{ISIC\_0032042} from testing and validation partitions\rev{,} respectively to the training partition.

\rev{For detecting duplicates in \ham based on image embedding similarity, we used \pypackage{fastdup} to extract a \featdim-dimensional embedding for all the \hamcount images. The similarities between all the $\binom{\hamcount}{2}$ pairs of images, denoted by $\mathcal{S}$, were calculated using the cosine similarity, where $0 \leq \mathcal{S} \leq 1$ and a higher cosine similarity score corresponds to a higher similarity.} It should be noted that using cosine similarity in a lower-dimensional embedding space yields more accurate matches, as compared to using it in the high-dimensional image space, where it results in \say{\textit{only false positive results}}, as noted by Cassidy et al.~\cite{cassidy2022analysis}.

\subsubsection{Image resizing}
The \lowres images in \dermaC and \dermaE were obtained by resizing the original high-resolution skin lesion images from \ham and other sources using bicubic interpolation. For the \hires resolution, unlike \dermamnist which first resized the images from the original resolution to \lowres using bicubic interpolation and then resizing to \hires using nearest neighbor interpolation, the images in \dermaC and \dermaE were obtained by directly resizing the original resolution images to \hires using bicubic interpolation. All image resizing operations were performed using the Python library \pypackage{PIL}.

\subsubsection{Model training and evaluation}
We used the official \medmnist training and evaluation code~\cite{medmnist_experiments}
by Yang et al.~\cite{yang2021medmnist,yang2023medmnist}. Both the model architectures: ResNet-18 and ResNet-50, are trained for 100 epochs with the cross-entropy loss and the Adam~\cite{kingma2014adam} optimizer with a batch size of 128. An initial learning rate of 0.001 was used, and a learning rate scheduler reduced it by a factor of 10 after the 50\textsuperscript{th} and the 75\textsuperscript{th} epochs. \ka{Over the training epochs,} the model with the best area under the ROC curve (AUC) on the validation partition was used for testing. The reported metrics for evaluation were the AUC and the overall classification accuracy (ACC). All models were trained and evaluated using PyTorch~\cite{paszke2019pytorch}.


\subsection{\fitzdset}

\subsubsection{Duplicate detection and manual verification}
\revnew{The images in \fitzdset were downloaded using the image URLs in the dataset's metadata.}
For all the \fitzcount{} images in \fitzdset, a \featdim-dimensional embedding was extracted using \pypackage{fastdup}. All subsequent analysis (duplicate pairs detection, duplicate clusters detection, outlier detection) used these embeddings. The similarities between all possible pairs of images (i.e., 
$\binom{\fitzcount{}}{2}$
possible pairs), denoted by $\mathcal{S}$, were calculated using the cosine similarity.

\begin{sloppypar}
The original filenames in \fitzdset follow the format \texttt{\textlangle MD5\_hash\textrangle.jpg}, which is arguably not helpful in interpreting either the diagnosis or the FST label without looking up the metadata. For an intuitive understanding of the images, we renamed them from their original filenames to a more interpretable format \texttt{\textlangle diag. abbrv.\textrangle\_f\textlangle FST label\textrangle\_\textlangle image index\textrangle\_\textlangle truncated MD5hash\textrangle.jpg}. Images with missing FST labels are assigned FST 0. For example, a file originally called \texttt{0a94359e7eaacd7178e06b2823777789.jpg} is renamed to \texttt{ps\_f1\_0\_0a94359e.jpg}, which can be interpreted as: this is the first image (index \texttt{0}) of \say{psoriasis} (\texttt{ps}) and has FST 1 (\texttt{f1}) and truncated MD5 hash \texttt{0a94359e}. We chose to include a truncated MD5 hash of length 8 to make it easier to map new filenames to the original ones while still avoiding hash collision.
\ka{A version of \fitzdset with renamed files} and the metadata containing all the old and new filenames and the diagnosis labels' abbreviations are available online~\cite{critique_metadata_cite}.
\end{sloppypar}

\begin{figure}[!ht]
\centering
\includegraphics[width=\textwidth]{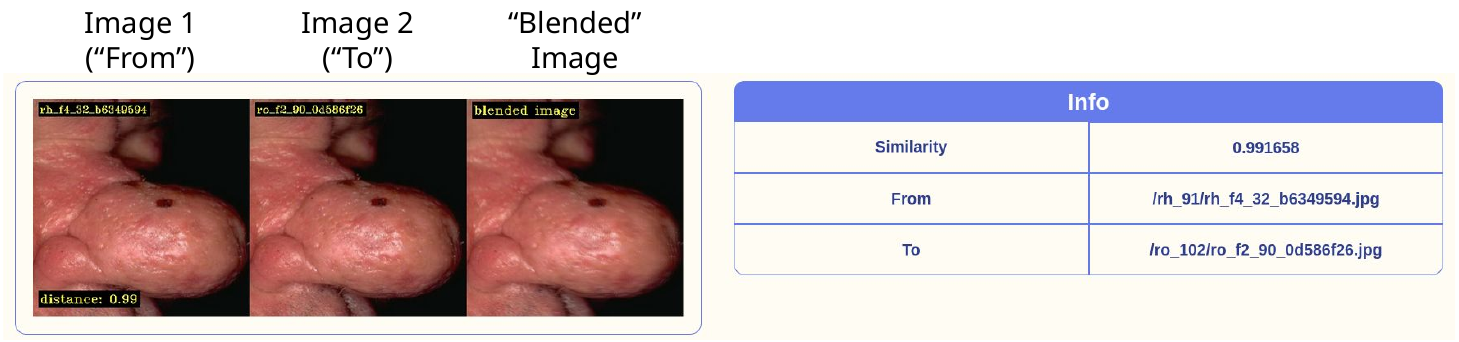}
\caption{Sample row from the \pypackage{fastdup} similarity detection results containing a pair of duplicate images along with their filenames and similarity score.}
\label{fig:understanding_fastdup}
\end{figure}

To detect duplicates, all image pairs' similarity scores were filtered based on the threshold(s), and this resulted in 6,622 unique image pairs with cosine similarity over 0.90 (i.e., $|\{\mathcal{S}_{0.90}\}|$) and 1,425 unique images pairs for with similarity over 0.95 (i.e., $|\{\mathcal{S}_{0.90}\}|$).
For the manual review of duplicates, a GUI was created using the Python library \pypackage{tkinter} that displayed the candidate duplicate image pairs detected by \pypackage{fastdup}, and three clickable buttons: \say{Duplicate}, \say{Unclear}, and \say{Different}. Two annotators were tasked with independently reviewing the 1,425 duplicate images pairs each, and their responses were recorded in a CSV file.

For using \pypackage{cleanvision} to detect duplicates, we filtered the image quality report for \say{near\_duplicates} and \say{exact\_duplicates}, which were 100 and 10 unique image pairs, respectively. Upon filtering these output pairs to those detected and verified using \pypackage{fastdup}'s output, we found 19 unique duplicate pairs that were not detected using \pypackage{fastdup}.

\subsubsection{Interpreting \pypackage{fastdup} duplicates visualization}

Fig.~\ref{fig:understanding_fastdup} shows a sample row from the table of duplicates in \fitzdset detected using \pypackage{fastdup} and available online~\cite{critique_duplicate_pairs_cite}.
This can be interpreted as:

\begin{itemize}
    \item {\texttt{/rh\_91/rh\_f4\_32\_b6349594.jpg} (Image 1)}: Image of \say{rhinophyma} (\texttt{rh}) with {FST 4}. This is image \#33 of total 91 rhinophyma images.
    \item {\texttt{/ro\_102/ro\_f2\_90\_0d586f26.jpg} (Image 2)}: Image of \say{rosacea} (\texttt{ro}) with {FST 2}. This is image \#91 of total 102 rosacea images.
    \item {\say{Blended} Image}: Output of $\alpha$-blending of the two images with $\alpha=0.5$.
    \item {Similarity}: The cosine similarity between the embeddings of the two images is 0.991658.
\end{itemize}

The \say{distance = 0.99} watermark on Image 1 is a misnomer. Despite being called \say{distance} (a \pypackage{fastdup} default), it is indeed reporting the cosine similarity value as expected.

\subsubsection{Erroneous image detection}
Using the image embeddings calculated for all the \fitzcount{} images, we adopt the following approach for detecting erroneous images: for each image $x_i$, we calculate its $N$ nearest neighbors $\{x_{i1}, \cdots, x_{iN}\}$ in the embedding space, and their corresponding similarities $\{\mathcal{S}_{i, i1}, \cdots, \mathcal{S}_{i, iN}\}$. An outlier would be dissimilar to other skin images in the dataset, and would therefore have low similarity scores with its nearest neighbors. To list all possible outliers in \fitzdset, we choose $N = 5$ and prepare tuples of $\left( x_i, min \left(\{\mathcal{S}_{i, i1}, \dots, \mathcal{S}_{i, i5}\}\right)  \right)$, where $min \left(\{\mathcal{S}_{i, i1}, \dots, \mathcal{S}_{i, i5}\}\right)$, called the outlier score, is inversely proportional to the likelihood of an image \rev{being} an erroneous image. These tuples are then sorted by the ascending order of their outlier score, i.e., the image with the lowest score and therefore the most likely to be an erroneous image is listed first, and are displayed online~\cite{critique_erroneous_images_cite}.

\subsubsection{\rev{Correcting} \fitzdset}
The \fitzdset cleaning pipeline consists of the following steps:
\begin{enumerate}
    \item \textbf{Similarity score-based filtering:} We specify an image similarity threshold of $0.99$, meaning any image whose maximum similarity score to any other image in the dataset is greater than this threshold is removed. This is done to remove the near-exact duplicates.
    \item \textbf{Processing duplicates:} We then process the duplicate pairs and clusters detected by \pypackage{fastdup} and \pypackage{cleanvision} to merge them into larger clusters if they exist \rev{(Fig.~\ref{fig:f17k_coalesce_duplicates})}, using the union-find algorithm~\cite{unionfind}. For the final list of clusters, we check if the clusters are \say{homogenous}, meaning if all the duplicate images in a cluster share the same diagnosis and FST labels. If a cluster is not homogenous, we remove all the images in it from \fitzdset, but if it is, we retain the single largest image in the cluster by spatial resolution and remove the rest.
    \item \textbf{Remove erroneous images}: Finally, we remove the erroneous images detected in \fitzdset.
\end{enumerate}

While we do not do this in \fitzdsetC, the data cleaning code also allows the users to remove images from \fitzdset that have missing FST labels.

\subsubsection{Model training and evaluation}
We use the official \fitzdset training and evaluation code~\cite{fitzpatrick17k_github}
by Groh et al.~\cite{groh2021evaluating} with some modifications. First, we use separate partitions for validation (i.e., picking the best performing model across training epochs based on 
the highest validation accuracy)
and for testing (i.e., for reporting the final \fitzdsetC benchmarks in Table~\ref{tab:f17k_results}), and therefore the code is modified accordingly. Next, 
\rev{Next, the datasets for the seven experimental settings (i.e., \say{Verified}, \say{Random}, \say{Source A}, \say{Source B}, \say{FST 3--6}, \say{FST 1--2 \& 5--6}, and \say{FST 1--4}) proposed by Groh et al.~\cite{groh2021evaluating} vary considerably in the number of images across training-validation-testing partitions (Table~\ref{tab:f17k_results}), we conduct a hyperparameter search for each setting. We vary the optimizer: \{Adam, SGD\}, the learning rate: \{$1e-2$, $1e-3$, $1e-4$\}, and the number of training epochs: \{20, 50, 100, 200\}, and for each of the 7 experimental settings, we train 3 models with each of the above hyperparameter settings with different seed values, effectively training $2 \times 3 \times 4 \times 7 \times 3 = 504$ models. For each experimental setting, the hyperparameter setting with the highest accuracy on the validation partition was used for the final testing, and the results are reported in Table~\ref{tab:f17k_results}.}
Finally, we used mixed precision training through Hugging Face Accelerate~\cite{hf_accelerate_cite_new}
to speed up the training. 
For all the experiments, 
we use
the same image transformations as those in Groh et al.'s work.
The reported metrics for evaluation were the overall and the FST-wise classification accuracies. All models were trained and evaluated using PyTorch~\cite{paszke2019pytorch} and Accelerate~\cite{hf_accelerate_cite_new}.

\rev{To understand how sensitive these classification models are to the hyperparameter choices, we also evaluated models optimized on one experiment's best hyperparameters on another experiment's test set, and these results are presented in Table~\ref{tab:f17k_hparam_7x7}. The columns represent the optimal hyperparameters for each setting, and the rows represent the overall test set accuracies for all the settings when evaluated using those particular hyperparameters. 
We observe that varying the hyperparameters does not considerably affect the test accuracies.
Additionally, the entries along the diagonal of Table~\ref{tab:f17k_hparam_7x7} are the same as the overall accuracies in Table~\ref{tab:f17k_results}, since these are the test accuracies of models trained and evaluated on each particular setting's optimal hyperparameters.}


\subsection{Hardware and Software Environments}
All experiments were carried out on a workstation running Ubuntu 20.04 with AMD Ryzen 9 16-core 5950X CPU, 32 GB RAM, and NVIDIA RTX 3090 24 GB GPU. The following versions of software packages were used: Python 3.10, \pypackage{torch} 1.11.0, \pypackage{torchvision} 0.12.0, \pypackage{PIL} 10.0.1, \pypackage{fastdup} 1.71, \pypackage{cleanvision} 0.3.4, and \pypackage{accelerate} 0.9.0.

\subsection{\rev{Source Datasets' Licenses}}

\rev{All the source datasets are associated with the Creative Commons (CC) Licenses: \dermamnist (CC BY-NC 4.0)~\cite{medmnist_website_cite},
\fitzdset (CC BY-NC-SA 3.0)~\cite{fitzpatrick17k_github},
and ISIC 2018 Challenge Datasets (CC BY-NC 4.0)~\cite{isic2018_website_cite}.
}

\section{Code availability}

\ka{The codebase for the analyses and the experiments in this paper is available at \url{\ghrepourl}. The results of analysis, including 
the visualizations, are available online at \url{https://derm.cs.sfu.ca/critique}.}

\section{Data availability}
\rev{All the datasets released with this work: \dermaC, \dermaE, and \fitzdsetC are publicly available on Zenodo~\cite{abhishek_2024_11101337}.
}

\section{Acknowledgements} 

K.A. is funded by Natural Sciences and Engineering Research Council of Canada (NSERC) Discovery grant RGPIN 06795 and Simon Fraser University. 
The authors would like to thank Rui Shi for sharing the image IDs of the training, validation, and testing partitions of the \dermamnist dataset. The authors are grateful to Prof. Rafeef Garbi for suggesting the idea of documenting and publicly disseminating these important data quality issues in a paper.
The authors are also thankful to Hanene Ben Yedder and Arezou Pakzad of the Medical Image Analysis Lab for the helpful discussions, and to the NVIDIA Corporation and the Digital Research Alliance of Canada for providing the computational resources.

\section{Author contributions statement}

\ka{K.A. worked on writing the code, performing the formal analysis and the experiments, and preparing the figures, with support from A.J. and G.H. K.A. and A.J. worked on data curation. K.A. worked on writing the initial draft. G.H. supervised the project and provided the resources and funding for the project. All authors contributed to the design and the evaluation of the algorithms. All authors contributed to writing, reviewing, and editing the manuscript. All authors read and approved the manuscript.}

\section{Competing interests} 

\ka{The authors declare no competing interests.}

\bibliography{sample,references}

\end{document}